\documentclass[10pt,twocolumn,letterpaper]{article}
\usepackage[pagenumbers]{cvpr}

\usepackage{graphicx}
\usepackage{booktabs}
\usepackage{multirow}
\usepackage{arydshln}
\usepackage{xcolor}
\usepackage{enumitem}
\usepackage{microtype}
\usepackage{algorithm}
\usepackage{algpseudocode}

\usepackage{amssymb,amsmath,amsfonts,epsfig,mdframed,rotating,cancel,mathtools}

\usepackage{bbm}

\usepackage{comment} 

\renewcommand{\[}{\left[}

\def\R{{\mathbb R}}

\newfont{\bbb}{msbm10 scaled 500}

\newfont{\bb}{msbm10 scaled 1100}

\newcommand{\ev}{{\bf e}}

\newcommand{\vv}{{\bf v}}
\newcommand{\xv}{{\bf x}}

\newcommand{\Em}{{\bf E}}

\newcommand{\Vm}{{\bf V}}

\newcommand{\Bc}{{\cal B}}
\newcommand{\Cc}{{\cal C}}
\newcommand{\Dc}{{\cal D}}

\newcommand{\Ic}{{\cal I}}

\newcommand{\Qc}{{\cal Q}}

\newcommand{\Sc}{{\cal S}}

\newcommand{\Uc}{{\cal U}}

\newcommand{\muv}{\hbox{\boldmath$\mu$}}

\newcommand{\Sigmam}{\hbox{\boldmath$\Sigma$}}

\renewcommand{\arg}{{\hbox{\rm arg}}}

\newcommand{\<}{\left\langle}
\renewcommand{\>}{\right\rangle}

\def\<{\langle}
\def\>{\rangle}

\newcommand{\I}{{\bf I}}

\usepackage{tikz,tkz-tab}

\usetikzlibrary{fit,automata,positioning,calc,shapes}
\usepackage{ifthen}

\ifdefined\circlenode
\else
\newcommand{\circlenode}[3]{
	\node [draw, circle, inner sep=2pt, #1] (#2) {#3};
}
\fi
\ifdefined\sqnode
\else
\newcommand{\sqnode}[3]{
	\node [draw, rectangle, inner sep=2pt, #1] (#2) {#3};
}
\fi

\definecolor{darkgreen}{rgb}{0.1,0.5,0}
\definecolor{Midnightblue}{rgb}{.2,.2,.7}
\definecolor{mypink3}{cmyk}{0, 0.7808, 0.4429, 0.1412}

\definecolor{amber}{rgb}{1.0, 0.49, 0.0}

\newcommand \remove[1]{ }

\definecolor{cvprblue}{rgb}{0.21,0.49,0.74}
\usepackage[breaklinks,colorlinks,allcolors=cvprblue]{hyperref}

\def\confName{CVPR}
\def\confYear{2027}

\title{GALA: Gallery-Level Alignment for Real-Life Vision Decoding from EEG}

\author{Philippe Forest$^{1}$ \qquad Giulia Lioi$^{1}$ \qquad Mai Quyen Pham$^{1}$\\ Alix Lamouroux$^{2}$ \qquad Adrien Merlini$^{1}$ \qquad Javen Shi$^{3}$\\[0.5em] {\small $^{1}$IMT Atlantique, France \qquad $^{2}$Center for Mind/Brain Sciences (CIMeC), University of Trento, Italy}\\ {\small $^{3}$Australian Institute for Machine Learning, University of Adelaide, Australia}}

\begin{document}
\maketitle

\begin{abstract}
Neural decoding of visual experience from electroencephalography (EEG) has achieved strong performance in intra-subject settings, but current methods fail to generalize across individuals: in cross-subject EEG-to-image retrieval on the standard 200-way benchmark THINGS-EEG2, top-1 accuracy drops from above 80\% within subjects to about 32\% for an unseen subject. Adapting a pre-trained encoder to new subjects and sessions typically requires labeled data, a bottleneck that limits practical deployment. In a practical brain-computer interface (BCI), however, the possible outcomes are known in advance and form a gallery of candidates, and the unlabeled EEG of the user accumulates as soon as the device is used. We propose a new setting, \emph{incremental transductive calibration}, a form of online test-time adaptation in which the decoder is calibrated regularly on all the unlabeled EEG received so far and on the known gallery, without any labels. It is closer to a deployed BCI than the zero-shot and batch evaluations in current use. In our protocol, queries arrive one at a time, with unknown and possibly unequal frequencies, and each query is decoded before it is used for calibration. We introduce \textbf{GALA} (\textbf{GA}llery-\textbf{L}evel \textbf{A}lignment), a label-free calibration method that registers the unlabeled EEG embeddings onto the gallery through shrinkage whitening, hubness correction and an optimal-transport (Sinkhorn--Procrustes) alignment with query frequencies estimated from the buffer, updated after every query. On THINGS-EEG2, with a gallery of 200 images, it raises the top-1 accuracy on a new subject from 28\% to 67\% over 53 minutes of use. Across 60 evaluation settings that vary the gallery size, the query frequencies and the number of repetitions averaged per query, GALA is the most accurate label-free method in every setting, 6.6 points above the best competing method on average at the end of the stream.
\end{abstract}

\begin{figure}[t]
  \centering
  \includegraphics[width=\columnwidth]{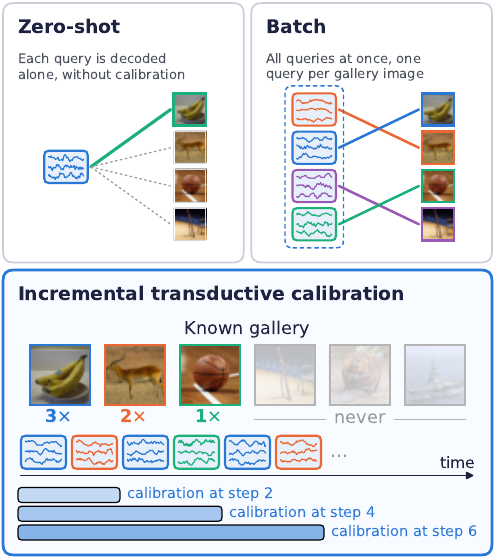}
  \caption{Evaluation settings. Top: in the standard zero-shot evaluation, each query is decoded alone against the gallery. In the standard transductive batch evaluation, the 200 test queries are calibrated together and each gallery image is queried exactly once. Bottom: in incremental transductive calibration, the gallery is known, but queries arrive one at a time with unequal frequencies, some gallery images are never queried, and the calibration at each step uses all the queries received so far.}
  \label{fig:protocol}
\end{figure}

\section{Introduction}
\label{sec:intro}

Decoding visual perception from brain activity is a central goal of neuroscience~\cite{kamitani2005decoding,carlson2013representational,grootswagers2017decoding}, with applications ranging from the study of visual representations~\cite{cichy2014resolving,contini2017decoding} to assistive brain-computer interfaces (BCIs) ~\cite{lawhern2018eegnet,gao2021interface}. EEG is especially attractive for this problem because it is non-invasive, has high temporal resolution, and is relatively inexpensive. In the now-standard \emph{EEG-to-image retrieval} setting~\cite{spampinato2017deep,gifford2022things,grootswagers2022things}, EEG signals and images are represented in a joint embedding space, and retrieval selects, for a given EEG signal, the closest image from a candidate set. Following progress in fMRI-to-image decoding~\cite{takagi2023highres,ozcelik2023natural,scotti2023mindeye,scotti2024mindeye2}, recent work aligns EEG embeddings with features from large pre-trained vision models through contrastive learning~\cite{song2024nice,li2024atm,wu2025ubp,zhang2025neurobridge,du2025shallow}, and now exceeds 80\% top-1 accuracy in the \emph{intra-subject} 200-way setting~\cite{du2025shallow}. In the \emph{cross-subject} setting, where the encoder must generalize to a held-out subject under the leave-one-subject-out (LOSO) protocol, accuracy remains between 20\% and 30\% for the strongest published encoders~\cite{du2025shallow,zhang2025neurobridge,cui2026score}.

This gap is rooted in the inter-subject variability typical of EEG: skull geometry, cortical folding and electrode placement induce a different spatial mixing of the underlying neural sources for each individual~\cite{onton2006imaging}, and inter-subject differences in covariance structure further amplify this mismatch~\cite{blankertz2008optimizing,barachant2010riemannian,samek2013transferring,haufe2014interpretation}. Beyond these anatomical differences, physiological factors (e.g., individual differences in brain activity and cognitive state) and experimental factors (e.g., recording conditions and equipment) further contribute to inter-subject variability.  How these differences affect generalization to new subjects remains an open question~\cite{li2026crosssubject,lin2026identitytrap}.

In a practical use of a brain-computer interface, the set of possible outcomes is known in advance. For example, a person who cannot communicate verbally may use a BCI to select items displayed on a screen, such as food, drinks, or objects they wish to request. The assistive interface presents a predefined set of options, and decoding consists of matching the user's EEG signal to one of these candidates in the gallery, corresponding to the retrieval setting described above. The main obstacle to deployment is the shift of the EEG signal between subjects, and between sessions of the same subject. Previous methods correct it by fine-tuning on labeled data~\cite{kneeland2025enigma}, but acquiring labeled calibration data for each new utilization of the interface is often impractical and time-consuming. A deployed decoder, however, knows its gallery of possible outcomes, and receives unlabeled signal from its user as soon as it is used.

Current evaluations do not reflect this situation. The THINGS-EEG2 test set was designed for zero-shot evaluation: it contains 200 images, each paired with one EEG query averaged over 80 repetitions, and every query is decoded alone. Most recent label-free test-time adaptation methods~\cite{huang2026sattc,cui2026score} are evaluated in a very similar setting, in which the 200 queries are available at once and every gallery image is queried exactly once. We therefore propose a new evaluation setting, \emph{incremental transductive calibration}, which simulates the deployment of a BCI: as the user operates the interface, its unlabeled EEG accumulates, and at every step the decoder is calibrated on all the EEG received so far, together with the known gallery (Fig.~\ref{fig:protocol}).  Our contributions are summarized as follows:
\begin{enumerate}[leftmargin=1.3em,itemsep=1pt,topsep=2pt]
  \item We introduce incremental transductive calibration, a realistic setting in which the decoder is calibrated at every step on the growing buffer of unlabeled EEG and the known gallery, and an evaluation protocol for it, in which every query is decoded before it is used for calibration, with skewed query frequencies and partially queried galleries.

  \item We introduce GALA, which casts calibration as a point-cloud registration problem~\cite{besl1992method,grave2019wasserstein} regularized by the zero-shot decoder as a prior, estimates the query frequencies from the buffer and is updated after every query. Across 60 evaluation settings on THINGS-EEG2, it is the most accurate label-free method in every setting over the last 30\% of the stream, 6.6 points above the best competing method on average.
  \item We also evaluate GALA in the standard batch setting, in which it reaches 79.1\% top-1 cross-subject accuracy without any labeled data from the new subject, 15.8 points above a Hungarian assignment that uses only the one-to-one structure of this test.
\end{enumerate}

The full training and evaluation code, as well as the configuration files for every reported run and the trained checkpoints, will be released publicly.

\section{Related Work}
\label{sec:related}

\paragraph{EEG-to-image decoding.}
Visual neural decoding has progressed from early supervised, subject-specific classifiers on labeled EEG-image pairs~\cite{spampinato2017deep,du2023bravl} to a zero-shot retrieval paradigm enabled by large-scale serial visual presentation benchmarks~\cite{gifford2022things,grootswagers2022things}.
NICE~\cite{song2024nice} established the contrastive alignment framework, using an InfoNCE~\cite{oord2019cpc} objective to align TSConv EEG embeddings with frozen CLIP~\cite{radford2021clip} image features.
UBP~\cite{wu2025ubp} dynamically blurs image targets proportional to stimulus uncertainty, while NeuroBridge~\cite{zhang2025neurobridge} introduced principled image augmentation for better bidirectional alignment.
ENIGMA~\cite{kneeland2025enigma} adapts to new subjects in a few-shot setting through subject-specific latent layers coupled with a compact bottleneck.
Shallow Alignment~\cite{du2025shallow} showed that intermediate-layer features of large pretrained vision models~\cite{chen2024internvl} better match the granularity of EEG, achieving 82.6\% intra-subject and 21.8\% cross-subject top-1 accuracy.

\paragraph{Cross-subject generalization.}
Inter-subject variability has long been identified as the central bottleneck for generalizable BCI systems~\cite{jayaram2016transfer,lotte2018review,samek2013transferring}. Training-time approaches such as adversarial inference~\cite{ozdenizci2020adversarial} and multi-source distribution matching~\cite{chen2021msmda,she2023multisource} improve robustness for paradigm-specific encoders, but do not provide a label-free correction at test time. EEG foundation models~\cite{elouahidi2025reve,cui2023neurogpt} aim at transferable representations, but do not explicitly address the alignment between neural signals and a visual embedding space. Riemannian re-centering~\cite{zanini2018transfer}, Euclidean alignment~\cite{he2020euclidean} and Riemannian Procrustes analysis~\cite{rodrigues2019rpa} align the covariance structure of a new subject without labels. These methods align the target to itself or to source statistics, ignoring the fact that in  a retrieval setting, the gallery provides a second, fixed point cloud in the embedding space. 

\paragraph{Label-free calibration for EEG-to-image retrieval.}
More recent work proposes label-free calibration methods for retrieval, such as 
SATTC~\cite{huang2026sattc}, which combines subject-adaptive whitening with adaptive local scaling and a structural expert, and reports an improvement from 9.2\% to 14.8\% top-1. SCORE~\cite{cui2026score} observes that subjects preserve concept relationships along different coordinate directions, trains the encoder on simulated recovery episodes, and at deployment estimates one orthogonal map from hubness-corrected mutual nearest neighbours, reaching 53.2\%. Both are transductive in the target data, and are evaluated on the 200-image test set, with all queries available at once and one query per image.

\paragraph{Transductive inference and test-time adaptation.}
Test-time adaptation updates normalization statistics or parameters on unlabeled target batches~\cite{li2018adabn,wang2021tent,liang2020shot,iwasawa2021t3a}, and transductive vision-language methods exploit the joint structure of a query batch~\cite{martin2024transductive,zanella2024transclip}. In the past, evaluations of these methods have repeatedly shown that gains measured on curated test batches do not survive realistic class imbalance, small batches or model selection without target labels~\cite{veilleux2021realistic,zhao2023pitfalls}. We apply these lessons to brain decoding.

\paragraph{Unsupervised adaptation during use.}
Adapting a decoder on the unlabeled data of its user has a long history in BCI: event-related potential spellers can be trained during use, without a calibration session, from the constraints of the application~\cite{kindermans2012bayesian} or from known label proportions~\cite{huebner2017learning}. In machine learning, online test-time adaptation updates a model on a stream of unlabeled inputs~\cite{wang2022cotta,gong2022note}, and label-shift methods estimate unknown class frequencies from the predictions of a fixed classifier~\cite{lipton2018bbse}. The setting we study differs from these lines of work in four respects: the task is retrieval over a gallery of embeddings that is known in advance, the calibration is estimated from a growing buffer, each query is decoded before it enters this buffer, and the query frequencies are unknown and may leave a part of the gallery unqueried.

\paragraph{Hubness and transductive alignment.}
Hubness~\cite{radovanovic2010hubs}, the tendency of a small subset of target vectors to dominate nearest-neighbor lists in high-dimensional spaces, is exacerbated when mapping source vectors between different domains~\cite{dinu2015hubness}. Cross-domain Similarity Local Scaling (CSLS)~\cite{conneau2018muse} addresses this issue by penalizing frequently-retrieved points, and is one of the components of SATTC~\cite{huang2026sattc}.
In unsupervised word translation, Conneau et al.~\cite{conneau2018muse} combine adversarial initialization with orthogonal Procrustes refinement~\cite{schonemann1966procrustes}, Artetxe et al.~\cite{artetxe2018robust} iterate CSLS retrieval and orthogonal mapping in a self-learning loop, Grave et al.~\cite{grave2019wasserstein} cast the problem as Wasserstein Procrustes~\cite{peyre2019computational} using Sinkhorn correspondence estimation~\cite{cuturi2013sinkhorn}, and Alvarez-Melis and Jaakkola~\cite{alvarezmelis2018gromov} use Gromov-Wasserstein distances. Functional alignment of brain responses across subjects follows the same principle with labeled, time-locked stimuli~\cite{haxby2011common,chen2015srm,scotti2024mindeye2}. We follow the same line of work, but transpose it to EEG-to-image retrieval, where the anchor of the alignment is the fixed image set, and the cloud of source data that we want to align grows during use.

\section{Incremental Transductive Calibration}
\label{sec:setting}

\paragraph{Problem formulation.}
We consider a multi-subject EEG-to-image decoding setting. Let $\xv_{s,i} \in \R^{C\times T}$ denote the EEG signal recorded from subject $s \in \Sc$ in response to image $i \in \Ic$, where $C$ is the number of EEG channels and $T$ the number of time steps, and let $\vv_i \in \R^{d}$ denote the corresponding image embedding. We observe a dataset $\Dc = \{(\xv_{s,i}, \vv_i): s\in \Sc, i \in \Ic\}$ where multiple subjects provide distinct EEG responses to the same visual stimulus. An encoder $E_\theta : \R^{C \times T} \to \R^d$ is trained on the source subjects $\Sc$ and then frozen. At test time, a new subject $s^\star \notin \Sc$ views images from a known gallery $\Vm = \{\vv_1,\ldots,\vv_K\}$, and the decoder receives unlabeled queries $\ev_t = E_\theta(\xv_{s^\star,i_t})$, $t=1,2,\ldots$, whose stimulus indices $i_t$ are unknown. At step $t$, the decoder holds the buffer $\Qc_t = \{\ev_1,\ldots,\ev_t\}$ of all the queries received so far. An \emph{incremental transductive calibration} is a sequence of corrections $c_t = \Cc(\Qc_t;\Vm)$, each computed from the whole buffer and the gallery only. The next query $\ev_{t+1}$ is decoded with $c_t$, before it enters the buffer. The correction must be defined for every buffer size: for an empty buffer it is the identity, which gives the zero-shot decoder, and at the start of a session it is estimated from few queries in a $d$-dimensional space. Calibration is specific to a subject and session. It requires no label and no dedicated calibration phase, since it uses the EEG that the user produces while naturally operating the interface.

\paragraph{Assumptions.}
The setting relies on four assumptions:
\begin{itemize}
    \item \textbf{(A1) Known gallery:} the set of candidates is known in advance. Zero-shot retrieval already makes this assumption, since the gallery is needed to rank the candidates.
    \item \textbf{(A2) Growing unlabeled buffer:} the queries of the user are stored, and the calibration at step $t$ uses all of them.
    \item \textbf{(A3) Unknown query frequencies:} the user does not query every gallery image equally often, and the frequencies are unknown to the decoder.
    \item \textbf{(A4) Partial coverage:} at any step, and possibly over the whole session, some gallery images have never been queried.
\end{itemize}

\paragraph{Standard evaluations.}
Prior work on THINGS-EEG2 uses two evaluations (Fig.~\ref{fig:protocol}). In the \emph{zero-shot} evaluation, each of the 200 test queries is decoded alone. In the \emph{batch} evaluation, the 200 queries are calibrated together, and each gallery image is queried exactly once. The batch evaluation is the limiting case of our setting in which the buffer is complete, with a bijection between queries and images in place of (A3) and (A4). This bijection alone is informative: a one-to-one (Hungarian) assignment~\cite{kuhn1955hungarian} of the zero-shot scores already raises accuracy substantially (Sec.~\ref{sec:conventional}).

\paragraph{Evaluation protocol.}
We simulate a session of use from the repetitions of the THINGS-EEG2 test set. A query is the average of $q$ repetitions of one gallery image, such that the 80 repetitions of an image provide $80/q$ disjoint queries. We consider three streams, which differ in how often each gallery image is queried. In the \emph{uniform} stream, every image is queried $80/q$ times. In the \emph{skewed} stream, the number of queries per image decreases exponentially from $80/q$ to 1. In the \emph{partial} stream, a random quarter of the gallery is never queried, and the number of queries per image decreases exponentially from $80/q$ to 1 over the other images. Queries arrive one at a time in random order, which mixes the four recording sessions. Every query is decoded once, by the calibration fitted on the queries received before it, and the calibration is updated after every query~\cite{dawid1984prequential}. The first five queries are decoded by the zero-shot decoder. The metric is the top-1 accuracy of these decisions, reported against the stimulus time elapsed. Methods that only decode the batch they are fitted on (SATTC, TransCLIP, EM-Dirichlet and the Hungarian assignment) are fitted on the buffer extended with the incoming query, and we keep their decision for this query. 

In this setting, the gallery $\Vm$ is a fixed point cloud in the embedding space, and the buffer of the new subject forms a second, growing cloud. If the encoder places each stimulus at a consistent position for every subject, up to a subject-specific transformation, the query cloud is a distorted copy of the queried part of the gallery. Calibration then amounts to registering the query cloud onto the gallery without known correspondences. Sec.~\ref{sec:method} describes how GALA estimates this registration.

\section{Gallery-Level Alignment}
\label{sec:method}

\begin{figure}[t]
  \centering
  \includegraphics[width=\columnwidth]{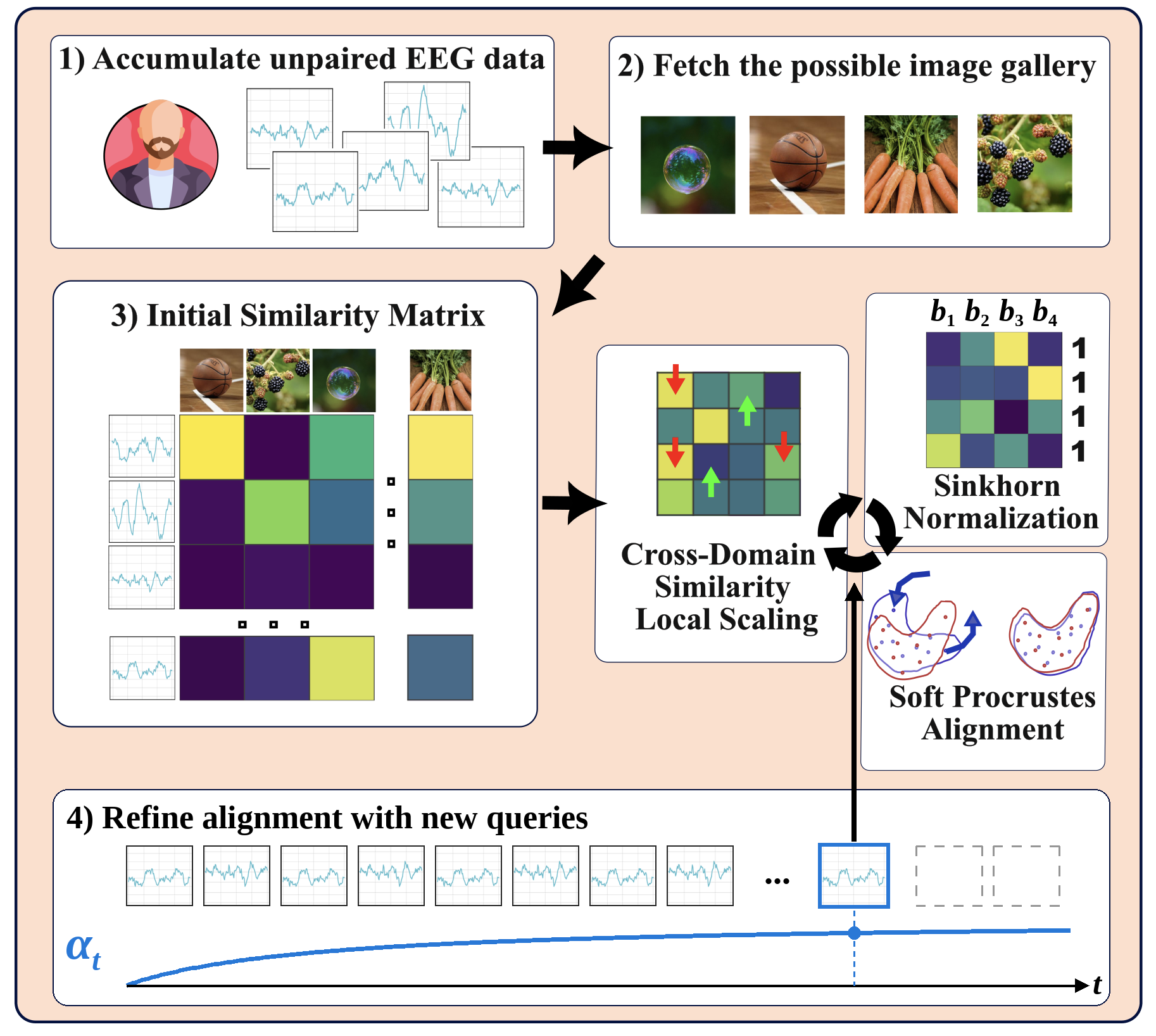}
  \caption{GALA. The unlabeled EEG embeddings of an unseen subject are whitened, compared with the gallery through cross-domain similarity local scaling, and registered onto it by a Sinkhorn normalization with estimated column masses $b_j$ and a soft Procrustes alignment. Each new query (blue) triggers one pass of this loop.}
  \label{fig:pipeline}
\end{figure}

At step $t$ of a session, the decoder holds the buffer  
 $\Em = [\ev_1^\top,\ldots,\ev_N^\top]^\top \in \R^{N\times d}$ of all the EEG embeddings received so far, with $N=t$, and the gallery $\Vm = [\vv_1^\top,\ldots,\vv_K^\top]^\top \in \R^{K\times d}$, the embedding of the $K$ candidate images, which are known in advance (A1). The gallery is fixed and does not depend on the user, whereas the buffer grows during the session, and the correspondence between queries and gallery images is unknown. GALA registers the buffer onto the gallery.  
\subsection{A prior for incremental calibration}
\label{sec:prior}
In the incremental setting, the calibration is updated after every query (A2), including at the start of a session, when the
buffer holds only a few queries in a $d$-dimensional space. Every quantity
estimated from the buffer is then unreliable, and a calibration that trusts
the buffer fully risks making the decoder worse than no calibration.
GALA is therefore built on one principle: the zero-shot decoder is a prior
worth $n_0$ observations, and the whitening and the rotation estimated from the
buffer are combined with their zero-shot values, with a weight on the buffer of
\begin{equation}
  \alpha_t = \frac{t}{t + n_0}.
  \label{eq:alpha}
\end{equation}
This prior enters the registration below: the moments used
for whitening shrink towards those of an isotropic cloud centered at the
origin, and the rotation is pulled towards the identity.
At $t = 0$, the whitening and the rotation are the identity, and GALA decodes
the zero-shot embeddings; as $t$ grows, the weight of
the prior vanishes and GALA converges to the unregularized registration.

\subsection{Whitening and registration}
\label{sec:problem}

As discussed in Sec.~\ref{sec:setting}, the buffer behaves as a distorted copy of the queried
part of the gallery. We model this distortion as a translation and an
anisotropic scaling, which whitening removes, followed by a change of
orientation, which the registration estimates jointly with the unknown
correspondences.
\paragraph{Whitening.}
Let $\muv$ and $\Sigmam$ be the mean and covariance of the buffer,
\begin{equation} \label{eq:moments}
\muv = \frac{1}{N}\sum_{i=1}^{N} \ev_i,
\qquad
\Sigmam = \frac{1}{N}\sum_{i=1}^{N}(\ev_i-\muv)(\ev_i-\muv)^\top .
\end{equation}
Following the prior of Sec.~\ref{sec:prior}, the mean is shrunk towards the origin and the covariance towards an isotropic one,
\begin{equation} \label{eq:shrunk}
\widetilde{\muv} = \alpha_t\,\muv,
\qquad
\widetilde{\Sigmam} = \alpha_t(1-\rho)\,\Sigmam + \big(1-\alpha_t(1-\rho)\big)\,\frac{\operatorname{tr}(\Sigmam)}{d}\, \I_d ,
\end{equation}
where $\rho$ is a shrinkage coefficient~\cite{ledoit2004honey}. The whitened and normalized EEG embeddings are
\begin{equation}\label{eq:whitening}
\widetilde{\ev}_i = \frac{\widetilde{\Sigmam}^{-1/2}(\ev_i-\widetilde{\muv})}{\|\widetilde{\Sigmam}^{-1/2}(\ev_i-\widetilde{\muv})\|}.
\end{equation}
We denote by $\widetilde{\Em}$ the matrix whose rows are $\widetilde{\ev}_i$. This step is the embedding-space counterpart of re-centering in classical BCI~\cite{zanini2018transfer,he2020euclidean}.
\paragraph{Registration problem.}
We estimate an orthogonal map $\mathbf{R}$, the query frequencies
$\mathbf{b}$ of the gallery images in the probability simplex $\Delta_K$, and a soft correspondence matrix
$\mathbf{P} \in \R_+^{t\times K}$ between queries and gallery images, by solving
\begin{subequations}\label{eq:problem}
    \begin{equation}
         \max_{\substack{\mathbf{R}^\top\mathbf{R} = \I_d,\ \mathbf{b} \in \Delta_K \\ \mathbf{P} \in \Pi(\mathbf{a},\mathbf{b})}}\;
   \big\langle \mathbf{P}, \mathbf{S}(\mathbf{R}) \big\rangle
  + \tau_{\mathrm{sk}}\, H(\mathbf{P})
  - \frac{n_0\,\bar\sigma}{2t}\,\big\|\mathbf{R} - \I_d\big\|^2 \label{eq:problema}
   \end{equation}
   \begin{equation}
         \text{s.t.}\quad  \mathbf{b} \in \arg\max_{\mathbf{b} \in \Delta_K}\, \ell(\mathbf{b}; \mathbf{R}),\label{eq:freq}
    \end{equation}
   \end{subequations} 
where $S_{ij}(\mathbf{R}) = \cos(\widetilde{\ev}_i\mathbf{R}, \vv_j)$ is the
similarity after registration, $H(\mathbf{P}) = -\sum_{ij}P_{ij}(\log P_{ij}-1)$
is the entropy of the correspondences, whose softness is controlled by the
temperature $\tau_{\mathrm{sk}}$, and
\begin{equation}
  \Pi(\mathbf{a},\mathbf{b}) = \big\{\mathbf{P} \ge 0 \,:\,
  \mathbf{P}\mathbf{1}_K = \mathbf{a},\ \mathbf{P}^\top\mathbf{1}_t = \mathbf{b}\big\}
  \label{eq:polytope}
\end{equation}
is the set of plans in which each query carries the mass
$\mathbf{a} = \tfrac1t\mathbf{1}_t$ and gallery image $j$ receives the mass
$b_j$. The last term is the prior of
Sec.~\ref{sec:prior} on the rotation; $\bar\sigma$, the mean singular value of
$\widetilde{\Em}^\top\mathbf{P}\Vm$, sets its scale. 

The masses $\mathbf{b}$ are the query frequencies, and estimating them lets GALA account for assumptions (A3) and (A4) of the incremental transductive calibration setting. GALA
therefore treats $\mathbf{b}$ as a variable, estimated from the buffer through the
constraint of~\eqref{eq:problem}. Each query is assigned to the gallery
independently of the others,
$\pi_{ij}(\mathbf{R}) = \exp(S_{ij}(\mathbf{R})/\tau_{\mathrm{sk}}) / \sum_{l}\exp(S_{il}(\mathbf{R})/\tau_{\mathrm{sk}})$,
and $\ell$ is the log-likelihood of these soft assignments under the
frequencies $\mathbf{b}$, with a symmetric Dirichlet prior worth $n_b$ queries per
image,
\begin{equation*}
  \ell(\mathbf{b};\mathbf{R}) =
  \sum_{j=1}^{K}\left(\sum_{i=1}^{t}\pi_{ij}
  \;+\; n_b \right)\log b_j
  \;=\; \sum_{j=1}^{K} w_j \log b_j.
\end{equation*}
Since all vectors have unit norm and $\mathbf{R}$ is orthogonal,
$\langle \mathbf{P}, \mathbf{S}(\mathbf{R})\rangle = \operatorname{tr}(\mathbf{R}^\top\widetilde{\Em}^\top\mathbf{P}\Vm)
= 1 - \tfrac12\sum_{i,j}P_{ij}\|\widetilde{\ev}_i\mathbf{R} - \vv_j\|^2$, so that without the prior term,
problem~\eqref{eq:problem} is the entropic Wasserstein--Procrustes problem~\cite{grave2019wasserstein}, which
minimizes the transport cost between the two clouds.

\begin{figure*}[t]
  \centering
  \includegraphics[width=\textwidth]{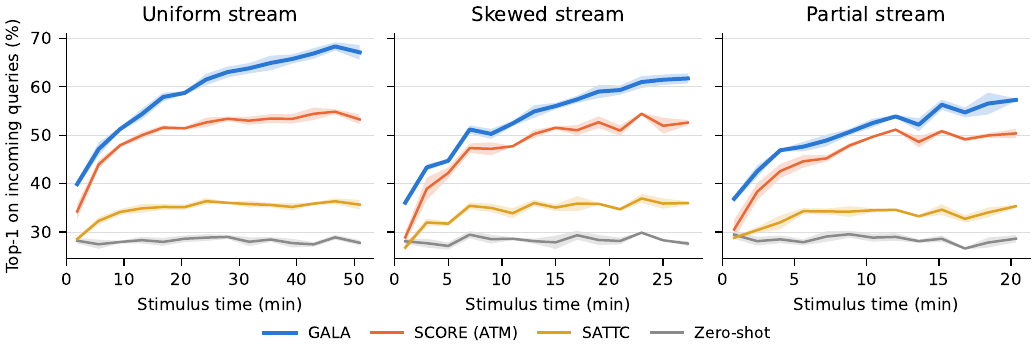}
  \caption{Incremental transductive calibration on THINGS-EEG2 (10 subjects, $q=20$). Top-1 accuracy of the incoming queries, each decoded before it is used for calibration, in windows of one fifteenth of the stream, against stimulus time, for the uniform, skewed and partial streams (mean $\pm$ one standard deviation over three training seeds). The figure shows four methods, all with the ATM encoder; Table~\ref{tab:inc} reports every method.}
  \label{fig:inc}
\end{figure*}

\subsection{Incremental update}
\label{sec:algo}

For fixed frequencies $\mathbf{b}$, problem~\eqref{eq:problem} is not jointly concave, but each block has an exact solution when the other is fixed, and we
solve it by alternating maximization. After
each query, GALA performs a single pass, starting from the rotation of the
previous step: it first updates the frequencies $\mathbf{b}$, then the
correspondences and the rotation (Alg.~\ref{alg:gala}).
\begin{algorithm}[ht!]
\caption{Update of GALA after query $t$.}
\label{alg:gala}
\begin{algorithmic}[1]
\Require buffer $\Em \in \R^{t\times d}$, gallery $\Vm \in \R^{K\times d}$, previous rotation $\mathbf{R}_{t-1}$ ($\mathbf{R}_0 = \mathbf{I}_d$);
         $n_0$, $n_b$, $\rho$, $\tau_{\mathrm{sk}}$, $\gamma$, $k$, $\mathbf{a}$
\State $\alpha_t \gets t/(t+n_0)$
\State $\widetilde{\Em} \gets$ whitened and normalized queries, Eqs.~\eqref{eq:moments}--\eqref{eq:whitening}
\State $\widehat{\mathbf{S}} \gets$ CSLS scores of $\widetilde{\Em}\mathbf{R}_{t-1}$ and $\Vm$, Eq.~\eqref{eq:csls}
\State $\mathbf{b} \gets$ estimated query frequencies, Eq.~\eqref{eq:masses}
\State $\mathbf{P} \gets \operatorname{Sinkhorn}\big(\exp(\widehat{\mathbf{S}}/\tau_{\mathrm{sk}});\,\mathbf{a},\mathbf{b}\big)$
\State $\mathbf{R}_t \gets \mathbf{U}\mathbf{W}^\top$, with $\mathbf{U}\mathbf{D}\mathbf{W}^\top$ the SVD of $\mathbf{M}$ in Eq.~\eqref{eq:M}, using $\mathbf{P}^{\circ\gamma}$
\State \Return the whitening and $\mathbf{R}_t$: query $t+1$ is decoded by $\arg\max_j$ of its CSLS scores after whitening and rotation
\end{algorithmic}
\end{algorithm}

\paragraph{Query frequencies.} For fixed $\mathbf{R}$, the frequencies are estimated
from the buffer. Maximizing~\eqref{eq:freq} under the constraint
$\sum_j b_j = 1$, with a Lagrange multiplier, gives $b_j = w_j / \sum_l w_l$. Since each row of $\boldsymbol{\pi}$
sums to one, $\sum_l w_l = K n_b + t$, and the mass of image $j$ is its soft
number of queries, combined with the prior of $n_b$ queries per image:
\begin{equation}
  b_j = \frac{n_b + \sum_{i=1}^{t}\pi_{ij}}{K n_b + t}.
  \label{eq:masses}
\end{equation}
At the start of a session, the masses are thus uniform, and they follow the
frequencies observed in the buffer as it grows.

\paragraph{Correspondence step.} For fixed $\mathbf{R}$ and $\mathbf{b}$, the optimal plan is
$\mathbf{P} = \operatorname{diag}(\mathbf{f})\exp(\mathbf{S}/\tau_{\mathrm{sk}})\operatorname{diag}(\mathbf{g})$,
where the scaling vectors $\mathbf{f} \in \R_+^N$ and $\mathbf{g} \in \R_+^K$ are computed by Sinkhorn
iterations~\cite{sinkhorn1964,cuturi2013sinkhorn} so that
$\mathbf{P} \in \Pi(\mathbf{a},\mathbf{b})$.

\paragraph{Rotation step.} For fixed $\mathbf{P}$, using the trace form of the alignment term and
$\|\mathbf{R}-\mathbf{I}_d\|_F^2 = 2d - 2\operatorname{tr}(\mathbf{R})$ for an orthogonal
$\mathbf{R}$, problem~\eqref{eq:problem} reduces to
$\max_{\mathbf{R}^\top\mathbf{R} = \mathbf{I}_d}\operatorname{tr}\big(\mathbf{R}^\top\mathbf{M}\big)$ with
\begin{equation}
  \mathbf{M} = \widetilde{\Em}^\top\mathbf{P}\,\Vm + \frac{n_0}{t}\,\bar\sigma\,\mathbf{I}_d ,
  \label{eq:M}
\end{equation}
an orthogonal Procrustes problem whose solution is $\mathbf{R} = \mathbf{U}\mathbf{W}^\top$,
where $\mathbf{U}\mathbf{D}\mathbf{W}^\top$ is the singular value decomposition of
$\mathbf{M}$~\cite{schonemann1966procrustes}. The prior term only adds
$\tfrac{n_0}{t}\bar\sigma\,\mathbf{I}_d$ to the data matrix: in directions where
the buffer carries little evidence, $\mathbf{M}$ is dominated by this term and
$\mathbf{R}$ stays close to the identity.

\paragraph{Practical choices.}
GALA departs from the exact block updates in two ways, both validated in Sec.~\ref{sec:ablation}.
First, since nearest-neighbour retrieval in high dimension is biased towards
hub images~\cite{radovanovic2010hubs}, the similarity $\mathbf{S}$ is replaced by the
Cross-domain Similarity Local Scaling score~\cite{conneau2018muse},
\begin{equation}
  \widehat{S}_{ij} = 2\cos(\widetilde{\ev}_i\mathbf{R}, \vv_j)
  - r_V(\widetilde{\ev}_i\mathbf{R}) - r_E(\vv_j),
  \label{eq:csls}
\end{equation}
where $r_V(\mathbf{x})$ is the mean cosine similarity of a query $\mathbf{x}$ to
its $k$ nearest gallery images, and $r_E(\vv_j)$ that of image $\vv_j$ to its $k$
nearest queries of the buffer. Second, the rotation step uses the sharpened plan
$\mathbf{P}^{\circ\gamma}$ in~\eqref{eq:M} and in $\bar\sigma$, the element-wise power with
$\gamma \ge 1$, which gives more weight to confident correspondences, and the
gallery embeddings in $\mathbf{M}$ are not normalized. An
incoming query is whitened, rotated and decoded with its CSLS scores,
$\hat\jmath = \arg\max_j \widehat{S}_{ij}$ (Alg.~\ref{alg:gala}).

\section{Experiments}
\label{sec:exp}

\subsection{Dataset and Experimental Setup}
\label{sec:dataset}

\begin{table*}[t]
\caption{Incremental transductive calibration: top-1 accuracy (\%) of the queries received after 1, 5 and 20 minutes of stimulus time, and at the end of the uniform stream. Every query is decoded before it is used for calibration ($q=20$, mean over 10 subjects, three training seeds and ten stream orders; one seed for SCORE with its own encoder).}
\label{tab:inc}
\centering
\small
\setlength{\tabcolsep}{5pt}
\begin{tabular}{@{}lcccccccccc@{}}
\toprule
 & \multicolumn{4}{c}{Uniform} & \multicolumn{3}{c}{Skewed} & \multicolumn{3}{c}{Partial} \\
\cmidrule(lr){2-5}\cmidrule(lr){6-8}\cmidrule(lr){9-11}
Method & 1 min & 5 min & 20 min & 53 min & 1 min & 5 min & 20 min & 1 min & 5 min & 20 min \\
\midrule
Zero-shot & 30.1 & 27.8 & 28.4 & 27.9 & 29.5 & 27.4 & 28.7 & 28.8 & 28.2 & 28.0 \\
CSLS & 35.8 & 34.9 & 35.4 & 34.2 & 34.6 & 34.2 & 35.7 & 35.1 & 35.5 & 34.5 \\
AdaBN~\cite{li2018adabn} & 30.1 & 28.9 & 30.3 & 29.1 & 29.5 & 29.0 & 30.6 & 28.8 & 31.9 & 29.6 \\
Tent~\cite{wang2021tent} & 30.1 & 28.6 & 30.0 & 29.0 & 29.5 & 28.7 & 30.4 & 28.8 & 31.5 & 29.4 \\
SATTC~\cite{huang2026sattc} & 28.2 & 32.0 & 35.5 & 35.9 & 27.5 & 32.2 & 35.7 & 28.3 & 33.2 & 34.3 \\
TransCLIP~\cite{zanella2024transclip} & 27.5 & 33.3 & 39.5 & 42.3 & 24.4 & 34.1 & 40.0 & 30.2 & 34.3 & 39.3 \\
EM-Dirichlet~\cite{martin2024transductive} & 26.6 & 35.4 & 38.7 & 42.2 & 27.7 & 35.4 & 39.2 & 25.9 & 35.1 & 37.9 \\
Hungarian assignment & 30.5 & 31.1 & 38.4 & 45.0 & 29.4 & 30.2 & 38.4 & 28.4 & 31.2 & 34.9 \\
SCORE~\cite{cui2026score} & 20.4 & 29.3 & 36.6 & 39.0 & 21.6 & 28.8 & 36.6 & 22.9 & 30.6 & 35.8 \\
SCORE with ATM & 29.3 & 42.3 & 52.0 & 53.4 & 29.3 & 42.2 & 52.1 & 29.9 & 43.9 & 50.1 \\
\midrule
GALA & \textbf{37.2} & \textbf{46.2} & \textbf{59.1} & \textbf{67.3} & \textbf{37.4} & \textbf{45.3} & \textbf{59.0} & \textbf{37.4} & \textbf{47.4} & \textbf{56.1} \\
\bottomrule
\end{tabular}
\end{table*}

THINGS-EEG2~\cite{gifford2022things} contains EEG recordings from 10 healthy participants who viewed images from 1,654 training-set object concepts and 200 testing-set concepts drawn from the THINGS database~\cite{hebart2019things}. Images are shown for 100 ms followed by a 100 ms blank screen. Each test concept is represented by one test image, recorded 80 times over four sessions. Our data preprocessing pipeline follows the methodology described in previous work~\cite{song2024nice,li2024atm,wu2025ubp}.

\paragraph{Evaluation.}
The evaluation task is \emph{200-way closed-set retrieval}: given the averaged EEG response of a held-out subject to one of the 200 test images, rank all 200 candidate images and report top-1 and top-5 accuracy (chance = 0.5\% and 2.5\% respectively). In the incremental setting, the gallery is the same 200 images, a query averages $q=20$ repetitions unless stated otherwise, and we follow the protocol of Sec.~\ref{sec:setting} with ten random stream orders per subject. With $q=20$, the uniform, skewed and partial streams last 53, 29 and 21 minutes of stimulus time. The accuracy after $m$ minutes is computed on the queries received between $0.7m$ and $1.3m$ minutes, and the accuracy at the end of a stream on its last 10\% of queries. We adopt the \emph{leave-one-subject-out} (LOSO) protocol: train on 9 subjects, evaluate on the 10th, and repeat for all 10 folds. Much of this line of work~\cite{zhang2025neurobridge,du2025shallow,huang2026sattc} selects the best checkpoint on the test-set loss of the held-out subject, which uses the test subject for model selection. To avoid this form of data leaking, we instead hold out 10\% of the training concepts for all 9 source subjects as a validation set, and select the checkpoint with the lowest contrastive loss on them (Sec.~\ref{sec:discussion}). 

\paragraph{Baselines.}
Among the label-free calibrations baselines, SATTC is evaluated with its official released code. SCORE has no released code at the time of submission, so we reproduce it from the paper, with its own encoder (SAMGA~\cite{jiang2026samga} with recovery-aware training) and its deployment step. This reproduction reaches 52.4\% in the batch evaluation, against 53.2\% reported (Sec.~\ref{sec:conventional}). Our experiments with SCORE have revealed that the method is considerably stronger when coupled with ATM~\cite{li2024atm} than with the SAMGA encoder originally used by the authors. For fairness, we also report SCORE in this stronger setting. As additional baselines, we add CSLS alone, the transductive vision-language methods EM-Dirichlet~\cite{martin2024transductive} and TransCLIP~\cite{zanella2024transclip}, with their official solvers. From test-time adaptation, we add AdaBN~\cite{li2018adabn} and Tent~\cite{wang2021tent}, which update the normalization layers of the encoder on the raw EEG of the buffer, every 20 queries. Finally, we add the \emph{Hungarian assignment}~\cite{kuhn1955hungarian} as a control, in which every gallery image receives the same number of queries: it uses the one-to-one structure of the batch test and no other information.

\subsection{Base encoder training}
\label{sec:encoder}

All calibrations act on frozen encoders trained with the same recipe: a multi-positive contrastive objective on stimulus-grouped batches~\cite{khosla2020supcon}, in which each trial is mixed with the trial of another source subject for the same stimulus. The objective is detailed in Appendix~\ref{app:encoder}.

\subsection{Implementation Details}
\label{sec:implementation}

Our method is implemented using PyTorch, and each model is trained on one NVIDIA GeForce RTX 3080 GPU. We employ a batch size of 1,024, AdamW with a learning rate of $3\times10^{-4}$, and train the models for 50 epochs (100 for ATM), with early stopping after 20 epochs without improvement of the validation loss. The temperature parameter $\tau$ is set to 0.07. Our main encoder is ATM~\cite{li2024atm} with a 128-dimensional embedding, trained with three seeds. All encoders are aligned to intermediate-layer (layer 28) features of InternViT-6B~\cite{chen2024internvl}, following Shallow Alignment~\cite{du2025shallow}. For Sec.~\ref{sec:encoders}, we train four more encoders with one seed: TSConv~\cite{song2024nice}, EEGNet~\cite{lawhern2018eegnet}, EEGProject~\cite{du2025shallow} and EEG Conformer~\cite{song2023eegconformer}. Finally, we fully fine-tune LaBraM~\cite{jiang2024labram}, a pretrained EEG foundation model, with a contrastive and regression objective, following the protocol of AVDE~\cite{dai2026avde}.
We also train ATM with one seed against five other image targets to study GALA's robustness to different alignment objectives (Appendix~\ref{app:encoders}). We systematically use a CSLS neighborhood $k=3$, 12 Sinkhorn normalization steps, a pseudo-count $n_b=1$ and one iteration per update, and select the shrinkage coefficient $\rho$, the Sinkhorn temperature $\tau_{\mathrm{sk}}$ and the power $\gamma$, as well as the two hyper-parameters of SCORE with ATM, cross-fold over a grid on the streams of the other subjects. The prior is set to $n_0 = 2d$ observations for every subject and every stream. The calibration runs on a standard CPU: on one core, one update of GALA takes 15\,ms with a buffer of 400 queries and 24\,ms with 800 queries, about ten times less than 16 iterations from the identity, which reach the same accuracy within 0.3 points (Table~\ref{tab:ablation}).

\subsection{Main Results}
\label{sec:inc}

Figure~\ref{fig:inc} and Table~\ref{tab:inc} compare GALA to existing approaches in the incremental setting. GALA is the most accurate method at every reported time of the three streams. After one minute, when the buffer holds about 15 queries, GALA and CSLS are the two most accurate methods. After five minutes, SCORE with ATM is the second-best method, 3.9 points below GALA on the uniform stream. It then grows more slowly, while GALA keeps improving with the size of the buffer: GALA reaches 59.1\% on the uniform stream after 20 minutes, and 67.3\% at the end of the stream, 13.9 points above SCORE. SATTC, TransCLIP and EM-Dirichlet stay at or below 40\% after 20 minutes. Appendix~\ref{app:stream} reports the same comparison for other menu sizes and numbers of repetitions: across a total of 60 evaluation settings with different gallery sizes, numbers of repetitions and stream types, GALA is the most accurate method over the last 30\% of the stream in every setting, by 6.6 points on average (Fig.~\ref{fig:grid-heatmap}).

\subsection{Ablation Study}
\label{sec:ablation}

Table~\ref{tab:ablation} (Appendix~\ref{app:ablation}) shows the individual contribution of each GALA component. The rotation is the main component: without it, accuracy after 20 minutes drops from 59.1\% to 39.1\% on the uniform stream. Without the prior, GALA is far below the zero-shot decoder after one minute, at 10.7\% against 30.1\% on the uniform stream. Whitening and CSLS each bring 1.6 to 3.7 points after 20 minutes. For the column masses of the plan, we compare the estimated frequencies of GALA with uniform masses and with a plan without column constraint. After 20 minutes, the estimated frequencies are 0.5 to 1.1 points above both alternatives on the three streams. 

\subsection{Generalization to Other Encoders and Datasets}
\label{sec:encoders}

To assess the robustness of our framework, we run the incremental protocol on five other EEG encoder architectures (Appendix~\ref{app:enc-tables}). GALA is above SCORE for every encoder at every reported time, except for LaBraM after 10 and 20 minutes. As the size of the gain depends strongly on the encoder, we provide an analysis in Appendix~\ref{app:encoders} that relates it to the similarity between the representational geometries of the EEG and image embeddings. Finally, we also run the incremental protocol with ATM on Alljoined-1.6M~\cite{xu2025alljoined} and THINGS-MEG~\cite{hebart2023thingsdata} (Appendix~\ref{app:datasets}). On these datasets, the zero-shot accuracy is much weaker due to a low signal-to-noise ratio, and every label-free method, including GALA, stays within 4 points of it.

\subsection{Standard Evaluations}
\label{sec:conventional}

Table~\ref{tab:main} (Appendix~\ref{app:batch}) reports the standard zero-shot and batch evaluations on the 200-image test set, with one 80-repetition query per image. In this evaluation, the buffer is complete, and GALA runs without the prior, with 16 iterations from the identity. It raises top-1 accuracy from 32.2\% to 79.1\%, and is better than every other label-free method we evaluate for every subject.

\paragraph{Hungarian control.}
Decoding the zero-shot scores with a one-to-one assignment raises accuracy from 32.2\% to 63.3\%. The one-to-one structure of the evaluation framework thus accounts for 31 points of accuracy, and only SCORE with ATM and GALA are above the level of this control. This highlights the need for more comprehensive and realistic evaluation setups, which is the motivation behind the incremental setting we propose.

\section{Discussion and Conclusion}
\label{sec:discussion}

\paragraph{Evaluation practice.}
Cross-subject EEG-to-image decoding is currently evaluated in conditions that are far from deployment, either without adaptation or with the whole test set at once, and accuracy on unseen subjects remains low. Incremental transductive calibration evaluates decoders in conditions closer to deployment, and GALA raises cross-subject accuracy over existing label-free methods in all the evaluated settings. Our experiments also point to two further limitations of current practice. First, THINGS-EEG2 provides no validation set, and much of the literature~\cite{zhang2025neurobridge,du2025shallow,huang2026sattc} selects the checkpoint on the held-out subject, which inflates the zero-shot accuracy of ATM by 3.6 points. We select all checkpoints on held-out training concepts. Second, a Hungarian assignment that uses only the one-to-one structure of the standard test set raises accuracy from 32.2\% to 63.3\%, and results on this test set should be reported together with this control.

\paragraph{Limitations and future work.}
 GALA's advantage depends on the base decoder: with LaBraM, SCORE is up to 1.7 points more accurate than GALA between 5 and 20 minutes of use (Appendix~\ref{app:enc-tables}), and the gain is small on Alljoined-1.6M and THINGS-MEG, where the zero-shot accuracy is below 10\%. Our evaluation, following the benchmark it builds on, replays offline recordings of a rapid serial visual presentation paradigm, in which each query averages many repetitions of the same image. A deployed BCI rarely offers these conditions: the candidate set changes with the task of the user, and queries come from few repetitions. Evaluations that reproduce these conditions, with task-dependent candidate sets, closed-loop sessions and single-trial queries, would measure more realistically what transductive calibration brings in practice. 

\paragraph{Conclusion.}
We argue for gallery-level alignment as a way to deploy EEG decoders on new users without labeled calibration data, formalize incremental transductive calibration, the assumptions under which it should be evaluated, and introduce GALA. Across 60 evaluation settings, GALA is the most accurate label-free method over the last 30\% of the stream in every setting. We hope that these contributions will help put forward incremental transductive calibration as a more realistic and practical alternative to the current benchmarking methods and advance research towards real-world BCI deployment.

\section*{LLM Usage Statement}
Claude (Anthropic) was used as a coding agent and an aid in the redaction of the article. It wrote and ran part of the experiments, produced some figures, and drafted and edited parts of the text. All of its outputs have been extensively verified by the authors. 

{
    \small
    \bibliographystyle{ieeenat_fullname}
    \bibliography{references}

\begin{thebibliography}{84}
\providecommand{\natexlab}[1]{#1}
\providecommand{\url}[1]{\texttt{#1}}
\expandafter\ifx\csname urlstyle\endcsname\relax
  \providecommand{\doi}[1]{doi: #1}\else
  \providecommand{\doi}{doi: \begingroup \urlstyle{rm}\Url}\fi

\bibitem[Alvarez-Melis and Jaakkola(2018)]{alvarezmelis2018gromov}
David Alvarez-Melis and Tommi Jaakkola.
\newblock {G}romov-{W}asserstein alignment of word embedding spaces.
\newblock In \emph{Proceedings of the 2018 Conference on Empirical Methods in
  Natural Language Processing}, pages 1881--1890, 2018.

\bibitem[Artetxe et~al.(2018)Artetxe, Labaka, and Agirre]{artetxe2018robust}
Mikel Artetxe, Gorka Labaka, and Eneko Agirre.
\newblock A robust self-learning method for fully unsupervised cross-lingual
  mappings of word embeddings.
\newblock In \emph{Proceedings of the 56th Annual Meeting of the Association
  for Computational Linguistics (Volume 1: Long Papers)}, pages 789--798, 2018.

\bibitem[Barachant et~al.(2010)Barachant, Bonnet, Congedo, and
  Jutten]{barachant2010riemannian}
Alexandre Barachant, St\'ephane Bonnet, Marco Congedo, and Christian Jutten.
\newblock Common spatial pattern revisited by {R}iemannian geometry.
\newblock In \emph{Proceedings of the IEEE International Workshop on Multimedia
  Signal Processing}, pages 472--476, 2010.

\bibitem[Besl and McKay(1992)]{besl1992method}
Paul~J. Besl and Neil~D. McKay.
\newblock A method for registration of 3-{D} shapes.
\newblock \emph{IEEE Transactions on Pattern Analysis and Machine
  Intelligence}, 14\penalty0 (2):\penalty0 239--256, 1992.

\bibitem[Blankertz et~al.(2008)Blankertz, Tomioka, Lemm, Kawanabe, and
  M\"uller]{blankertz2008optimizing}
Benjamin Blankertz, Ryota Tomioka, Steven Lemm, Motoaki Kawanabe, and
  Klaus-Robert M\"uller.
\newblock Optimizing spatial filters for robust {EEG} single-trial analysis.
\newblock \emph{IEEE Signal Processing Magazine}, 25\penalty0 (1):\penalty0
  41--56, 2008.

\bibitem[Carlson et~al.(2013)Carlson, Tovar, Alink, and
  Kriegeskorte]{carlson2013representational}
Thomas Carlson, David~A. Tovar, Arjen Alink, and Nikolaus Kriegeskorte.
\newblock Representational dynamics of object vision: The first 1000 ms.
\newblock \emph{Journal of Vision}, 13\penalty0 (10):\penalty0 1, 2013.

\bibitem[Chen et~al.(2021)Chen, Jin, Li, Fan, Li, and He]{chen2021msmda}
Hao Chen, Ming Jin, Zhunan Li, Cunhang Fan, Jinpeng Li, and Huiguang He.
\newblock {MS-MDA}: Multisource marginal distribution adaptation for
  cross-subject and cross-session {EEG} emotion recognition.
\newblock \emph{Frontiers in Neuroscience}, 15:\penalty0 778488, 2021.

\bibitem[Chen et~al.(2015)Chen, Chen, Yeshurun, Hasson, Haxby, and
  Ramadge]{chen2015srm}
Po-Hsuan Chen, Janice Chen, Yaara Yeshurun, Uri Hasson, James~V. Haxby, and
  Peter~J. Ramadge.
\newblock A reduced-dimension {fMRI} shared response model.
\newblock In \emph{Advances in Neural Information Processing Systems
  ({NeurIPS})}, 2015.

\bibitem[Chen et~al.(2024)Chen, Wu, Wang, Su, Chen, Xing, Zhong, Zhang, Zhu,
  Lu, Li, Luo, Lu, Qiao, and Dai]{chen2024internvl}
Zhe Chen, Jiannan Wu, Wenhai Wang, Weijie Su, Guo Chen, Sen Xing, Muyan Zhong,
  Qinglong Zhang, Xizhou Zhu, Lewei Lu, Bin Li, Ping Luo, Tong Lu, Yu Qiao, and
  Jifeng Dai.
\newblock {InternVL}: Scaling up vision foundation models and aligning for
  generic visual-linguistic tasks.
\newblock In \emph{Proceedings of the IEEE/CVF Conference on Computer Vision
  and Pattern Recognition ({CVPR})}, pages 24185--24198, 2024.

\bibitem[Cherti et~al.(2023)Cherti, Beaumont, Wightman, Wortsman, Ilharco,
  Gordon, Schuhmann, Schmidt, and Jitsev]{cherti2023openclip}
Mehdi Cherti, Romain Beaumont, Ross Wightman, Mitchell Wortsman, Gabriel
  Ilharco, Cade Gordon, Christoph Schuhmann, Ludwig Schmidt, and Jenia Jitsev.
\newblock Reproducible scaling laws for contrastive language-image learning.
\newblock In \emph{Proceedings of the IEEE/CVF Conference on Computer Vision
  and Pattern Recognition ({CVPR})}, pages 2818--2829, 2023.

\bibitem[Cichy et~al.(2014)Cichy, Pantazis, and Oliva]{cichy2014resolving}
Radoslaw~Martin Cichy, Dimitrios Pantazis, and Aude Oliva.
\newblock Resolving human object recognition in space and time.
\newblock \emph{Nature Neuroscience}, 17\penalty0 (3):\penalty0 455--462, 2014.

\bibitem[Conneau et~al.(2018)Conneau, Lample, Ranzato, Denoyer, and
  J{\'e}gou]{conneau2018muse}
Alexis Conneau, Guillaume Lample, Marc'Aurelio Ranzato, Ludovic Denoyer, and
  Herv{\'e} J{\'e}gou.
\newblock Word translation without parallel data.
\newblock In \emph{Proceedings of the International Conference on Learning
  Representations ({ICLR})}, 2018.

\bibitem[Contini et~al.(2017)Contini, Wardle, and Carlson]{contini2017decoding}
Erika~W. Contini, Susan~G. Wardle, and Thomas~A. Carlson.
\newblock Decoding the time-course of object recognition in the human brain:
  From visual features to categorical decisions.
\newblock \emph{Neuropsychologia}, 105:\penalty0 165--176, 2017.

\bibitem[Cui et~al.(2024)Cui, Jeong, Th\"olke, Medani, Jerbi, Joshi, and
  Leahy]{cui2023neurogpt}
Wenhui Cui, Woojae Jeong, Philipp Th\"olke, Takfarinas Medani, Karim Jerbi,
  Anand~A. Joshi, and Richard~M. Leahy.
\newblock {Neuro-GPT}: Towards a foundation model for {EEG}.
\newblock In \emph{Proceedings of the IEEE International Symposium on
  Biomedical Imaging ({ISBI})}, pages 1--5, 2024.

\bibitem[Cui et~al.(2026)Cui, Kan, Li, Wang, and Wu]{cui2026score}
Zhenyao Cui, Siyuan Kan, Siyang Li, Ziwei Wang, and Dongrui Wu.
\newblock {SCORE}: Subject coordinate recovery for label-free cross-subject
  {EEG}-to-image retrieval.
\newblock \emph{arXiv preprint arXiv:2608.19134}, 2026.

\bibitem[Cuturi(2013)]{cuturi2013sinkhorn}
Marco Cuturi.
\newblock Sinkhorn distances: Lightspeed computation of optimal transport.
\newblock In \emph{Advances in Neural Information Processing Systems
  ({NeurIPS})}, pages 2292--2300, 2013.

\bibitem[Dai et~al.(2026)Dai, Xiao, Yu, and Ye]{dai2026avde}
Sicheng Dai, Hongwang Xiao, Shan Yu, and Qiwei Ye.
\newblock Autoregressive visual decoding from {EEG} signals.
\newblock In \emph{Proceedings of the International Conference on Learning
  Representations ({ICLR})}, 2026.

\bibitem[Dawid(1984)]{dawid1984prequential}
A.~P. Dawid.
\newblock Present position and potential developments: Some personal views:
  {S}tatistical theory: The prequential approach.
\newblock \emph{Journal of the Royal Statistical Society. Series A (General)},
  147\penalty0 (2):\penalty0 278--292, 1984.

\bibitem[Dinu et~al.(2015)Dinu, Lazaridou, and Baroni]{dinu2015hubness}
Georgiana Dinu, Angeliki Lazaridou, and Marco Baroni.
\newblock Improving zero-shot learning by mitigating the hubness problem.
\newblock In \emph{Proceedings of the International Conference on Learning
  Representations ({ICLR}) Workshop}, 2015.

\bibitem[Du et~al.(2023)Du, Fu, Li, and He]{du2023bravl}
Changde Du, Kaicheng Fu, Jinpeng Li, and Huiguang He.
\newblock Decoding visual neural representations by multimodal learning of
  brain-visual-linguistic features.
\newblock \emph{IEEE Transactions on Pattern Analysis and Machine
  Intelligence}, 45\penalty0 (9):\penalty0 10760--10777, 2023.

\bibitem[Du et~al.(2026)Du, Dai, Song, Thompson, Tang, and Zhan]{du2025shallow}
Yang Du, Siyuan Dai, Yonghao Song, Paul~M. Thompson, Haoteng Tang, and Liang
  Zhan.
\newblock Deep models, shallow alignment: Uncovering the granularity mismatch
  in neural decoding.
\newblock \emph{Transactions on Machine Learning Research}, 2026.

\bibitem[{El Ouahidi} et~al.(2025){El Ouahidi}, Lys, Th\"olke, Farrugia,
  Pasdeloup, Gripon, Jerbi, and Lioi]{elouahidi2025reve}
Yassine {El Ouahidi}, Jonathan Lys, Philipp Th\"olke, Nicolas Farrugia, Bastien
  Pasdeloup, Vincent Gripon, Karim Jerbi, and Giulia Lioi.
\newblock {REVE}: A foundation model for {EEG} -- adapting to any setup with
  large-scale pretraining on 25,000 subjects.
\newblock In \emph{Advances in Neural Information Processing Systems
  ({NeurIPS})}, pages 22541--22577, 2025.

\bibitem[Gao et~al.(2021)Gao, Wang, Chen, and Gao]{gao2021interface}
Xiaorong Gao, Yijun Wang, Xiaogang Chen, and Shangkai Gao.
\newblock Interface, interaction, and intelligence in generalized
  brain--computer interfaces.
\newblock \emph{Trends in Cognitive Sciences}, 25\penalty0 (8):\penalty0
  671--684, 2021.

\bibitem[Gifford et~al.(2022)Gifford, Dwivedi, Roig, and
  Cichy]{gifford2022things}
Alessandro~T. Gifford, Kshitij Dwivedi, Gemma Roig, and Radoslaw~M. Cichy.
\newblock A large and rich {EEG} dataset for modeling human visual object
  recognition.
\newblock \emph{NeuroImage}, 264:\penalty0 119754, 2022.

\bibitem[Gong et~al.(2022)Gong, Jeong, Kim, Kim, Shin, and Lee]{gong2022note}
Taesik Gong, Jongheon Jeong, Taewon Kim, Yewon Kim, Jinwoo Shin, and Sung-Ju
  Lee.
\newblock {NOTE}: Robust continual test-time adaptation against temporal
  correlation.
\newblock In \emph{Advances in Neural Information Processing Systems
  ({NeurIPS})}, pages 27253--27266, 2022.

\bibitem[Grave et~al.(2019)Grave, Joulin, and Berthet]{grave2019wasserstein}
Edouard Grave, Armand Joulin, and Quentin Berthet.
\newblock Unsupervised alignment of embeddings with {Wasserstein Procrustes}.
\newblock In \emph{Proceedings of the International Conference on Artificial
  Intelligence and Statistics ({AISTATS})}, pages 1880--1890, 2019.

\bibitem[Grootswagers et~al.(2017)Grootswagers, Wardle, and
  Carlson]{grootswagers2017decoding}
Tijl Grootswagers, Susan~G. Wardle, and Thomas~A. Carlson.
\newblock Decoding dynamic brain patterns from evoked responses: A tutorial on
  multivariate pattern analysis applied to time series neuroimaging data.
\newblock \emph{Journal of Cognitive Neuroscience}, 29\penalty0 (4):\penalty0
  677--697, 2017.

\bibitem[Grootswagers et~al.(2022)Grootswagers, Zhou, Robinson, Hebart, and
  Carlson]{grootswagers2022things}
Tijl Grootswagers, Ivy Zhou, Amanda~K. Robinson, Martin~N. Hebart, and
  Thomas~A. Carlson.
\newblock Human {EEG} recordings for 1,854 concepts presented in rapid serial
  visual presentation streams.
\newblock \emph{Scientific Data}, 9:\penalty0 3, 2022.

\bibitem[Haufe et~al.(2014)Haufe, Meinecke, G\"{o}rgen, D\"{a}hne, Haynes,
  Blankertz, and Bie\ss{}mann]{haufe2014interpretation}
Stefan Haufe, Frank Meinecke, Kai G\"{o}rgen, Sven D\"{a}hne, John-Dylan
  Haynes, Benjamin Blankertz, and Felix Bie\ss{}mann.
\newblock On the interpretation of weight vectors of linear models in
  multivariate neuroimaging.
\newblock \emph{NeuroImage}, 87:\penalty0 96--110, 2014.

\bibitem[Haxby et~al.(2011)Haxby, Guntupalli, Connolly, Halchenko, Conroy,
  Gobbini, Hanke, and Ramadge]{haxby2011common}
James~V. Haxby, J.~Swaroop Guntupalli, Andrew~C. Connolly, Yaroslav~O.
  Halchenko, Bryan~R. Conroy, M.~Ida Gobbini, Michael Hanke, and Peter~J.
  Ramadge.
\newblock A common, high-dimensional model of the representational space in
  human ventral temporal cortex.
\newblock \emph{Neuron}, 72\penalty0 (2):\penalty0 404--416, 2011.

\bibitem[He and Wu(2020)]{he2020euclidean}
He He and Dongrui Wu.
\newblock Transfer learning for brain--computer interfaces: A {E}uclidean space
  data alignment approach.
\newblock \emph{IEEE Transactions on Biomedical Engineering}, 67\penalty0
  (2):\penalty0 399--410, 2020.

\bibitem[Hebart et~al.(2019)Hebart, Dickter, Kidder, Kwok, Corriveau, {Van
  Wicklin}, and Baker]{hebart2019things}
Martin~N. Hebart, Adam~H. Dickter, Alexis Kidder, Wan~Y. Kwok, Anna Corriveau,
  Caitlin {Van Wicklin}, and Chris~I. Baker.
\newblock {THINGS}: A database of 1,854 object concepts and more than 26,000
  naturalistic object images.
\newblock \emph{PLOS ONE}, 14\penalty0 (10):\penalty0 e0223792, 2019.

\bibitem[Hebart et~al.(2023)Hebart, Contier, Teichmann, Rockter, Zheng, Kidder,
  Corriveau, Vaziri-Pashkam, and Baker]{hebart2023thingsdata}
Martin~N. Hebart, Oliver Contier, Lina Teichmann, Adam~H. Rockter, Charles~Y.
  Zheng, Alexis Kidder, Anna Corriveau, Maryam Vaziri-Pashkam, and Chris~I.
  Baker.
\newblock {THINGS-data}, a multimodal collection of large-scale datasets for
  investigating object representations in human brain and behavior.
\newblock \emph{eLife}, 12:\penalty0 e82580, 2023.

\bibitem[Huang and Zhu(2026)]{huang2026sattc}
Qunjie Huang and Weina Zhu.
\newblock {SATTC}: Structure-aware label-free test-time calibration for
  cross-subject {EEG}-to-image retrieval.
\newblock In \emph{Proceedings of the IEEE/CVF Conference on Computer Vision
  and Pattern Recognition ({CVPR})}, pages 16887--16896, 2026.

\bibitem[H{\"u}bner et~al.(2017)H{\"u}bner, Verhoeven, Schmid, M{\"u}ller,
  Tangermann, and Kindermans]{huebner2017learning}
David H{\"u}bner, Thibault Verhoeven, Konstantin Schmid, Klaus-Robert
  M{\"u}ller, Michael Tangermann, and Pieter-Jan Kindermans.
\newblock Learning from label proportions in brain-computer interfaces: Online
  unsupervised learning with guarantees.
\newblock \emph{PLoS ONE}, 12\penalty0 (4):\penalty0 e0175856, 2017.

\bibitem[Iwasawa and Matsuo(2021)]{iwasawa2021t3a}
Yusuke Iwasawa and Yutaka Matsuo.
\newblock Test-time classifier adjustment module for model-agnostic domain
  generalization.
\newblock In \emph{Advances in Neural Information Processing Systems
  ({NeurIPS})}, pages 2427--2440, 2021.

\bibitem[Jayaram et~al.(2016)Jayaram, Alamgir, Altun, Sch\"{o}lkopf, and
  Grosse-Wentrup]{jayaram2016transfer}
Vinay Jayaram, Morteza Alamgir, Yasemin Altun, Bernhard Sch\"{o}lkopf, and
  Moritz Grosse-Wentrup.
\newblock Transfer learning in brain-computer interfaces.
\newblock \emph{IEEE Computational Intelligence Magazine}, 11\penalty0
  (1):\penalty0 20--31, 2016.

\bibitem[Jiang et~al.(2026)Jiang, She, Xu, Xu, Wu, and Kuang]{jiang2026samga}
Lin Jiang, Qingshan She, Jiale Xu, Haiqi Xu, Duanpo Wu, and Zhenzhong Kuang.
\newblock Subject-aware multi-granularity alignment for zero-shot
  {EEG}-to-image retrieval.
\newblock \emph{arXiv preprint arXiv:2604.17782}, 2026.

\bibitem[Jiang et~al.(2024)Jiang, Zhao, and Lu]{jiang2024labram}
Wei-Bang Jiang, Li-Ming Zhao, and Bao-Liang Lu.
\newblock Large brain model for learning generic representations with
  tremendous {EEG} data in {BCI}.
\newblock In \emph{Proceedings of the International Conference on Learning
  Representations ({ICLR})}, 2024.

\bibitem[Kamitani and Tong(2005)]{kamitani2005decoding}
Yukiyasu Kamitani and Frank Tong.
\newblock Decoding the visual and subjective contents of the human brain.
\newblock \emph{Nature Neuroscience}, 8\penalty0 (5):\penalty0 679--685, 2005.

\bibitem[Khosla et~al.(2020)Khosla, Teterwak, Wang, Sarna, Tian, Isola,
  Maschinot, Liu, and Krishnan]{khosla2020supcon}
Prannay Khosla, Piotr Teterwak, Chen Wang, Aaron Sarna, Yonglong Tian, Phillip
  Isola, Aaron Maschinot, Ce Liu, and Dilip Krishnan.
\newblock Supervised contrastive learning.
\newblock In \emph{Advances in Neural Information Processing Systems
  ({NeurIPS})}, pages 18661--18673, 2020.

\bibitem[Kindermans et~al.(2012)Kindermans, Verstraeten, and
  Schrauwen]{kindermans2012bayesian}
Pieter-Jan Kindermans, David Verstraeten, and Benjamin Schrauwen.
\newblock A {B}ayesian model for exploiting application constraints to enable
  unsupervised training of a {P300}-based {BCI}.
\newblock \emph{PLoS ONE}, 7\penalty0 (4):\penalty0 e33758, 2012.

\bibitem[Kneeland et~al.(2026)Kneeland, Jiang, Nunes, Scotti, Delorme, and
  Xu]{kneeland2025enigma}
Reese Kneeland, Wangshu Jiang, Ugo~Bruzadin Nunes, Paul~Steven Scotti, Arnaud
  Delorme, and Jonathan Xu.
\newblock {ENIGMA}: {EEG}-to-image in 15 minutes using less than 1\% of the
  parameters.
\newblock \emph{Transactions on Machine Learning Research}, 2026.

\bibitem[Kriegeskorte et~al.(2008)Kriegeskorte, Mur, and
  Bandettini]{kriegeskorte2008rsa}
Nikolaus Kriegeskorte, Marieke Mur, and Peter~A. Bandettini.
\newblock Representational similarity analysis -- connecting the branches of
  systems neuroscience.
\newblock \emph{Frontiers in Systems Neuroscience}, 2:\penalty0 4, 2008.

\bibitem[Kuhn(1955)]{kuhn1955hungarian}
Harold~W. Kuhn.
\newblock The {H}ungarian method for the assignment problem.
\newblock \emph{Naval Research Logistics Quarterly}, 2\penalty0
  (1--2):\penalty0 83--97, 1955.

\bibitem[Lawhern et~al.(2018)Lawhern, Solon, Waytowich, Gordon, Hung, and
  Lance]{lawhern2018eegnet}
Vernon~J. Lawhern, Amelia~J. Solon, Nicholas~R. Waytowich, Stephen~M. Gordon,
  Chou~P. Hung, and Brent~J. Lance.
\newblock {EEGNet}: A compact convolutional neural network for {EEG}-based
  brain--computer interfaces.
\newblock \emph{Journal of Neural Engineering}, 15\penalty0 (5):\penalty0
  056013, 2018.

\bibitem[Ledoit and Wolf(2004)]{ledoit2004honey}
Olivier Ledoit and Michael Wolf.
\newblock A well-conditioned estimator for large-dimensional covariance
  matrices.
\newblock \emph{Journal of Multivariate Analysis}, 88\penalty0 (2):\penalty0
  365--411, 2004.

\bibitem[Li et~al.(2024)Li, Wei, Li, Zou, and Liu]{li2024atm}
Dongyang Li, Chen Wei, Shiying Li, Jiachen Zou, and Quanying Liu.
\newblock Visual decoding and reconstruction via {EEG} embeddings with guided
  diffusion.
\newblock In \emph{Advances in Neural Information Processing Systems
  ({NeurIPS})}, pages 102822--102864, 2024.

\bibitem[Li et~al.(2026)Li, Yan, Dou, Song, and Zhang]{li2026crosssubject}
Taida Li, Yujun Yan, Fei Dou, Wenzhan Song, and Xiang Zhang.
\newblock Cross-subject generalization for {EEG} decoding: A survey of deep
  learning methods.
\newblock \emph{Progress in Biomedical Engineering}, 8\penalty0 (2):\penalty0
  022013, 2026.

\bibitem[Li et~al.(2018)Li, Wang, Shi, Hou, and Liu]{li2018adabn}
Yanghao Li, Naiyan Wang, Jianping Shi, Xiaodi Hou, and Jiaying Liu.
\newblock Adaptive batch normalization for practical domain adaptation.
\newblock \emph{Pattern Recognition}, 80:\penalty0 109--117, 2018.

\bibitem[Li et~al.(2025)Li, Kang, Gong, Dong, Zeng, Yan, Siok, and
  Wang]{li2024mcrl}
Yueyang Li, Zijian Kang, Shengyu Gong, Wenhao Dong, Weiming Zeng, Hongjie Yan,
  Wai~Ting Siok, and Nizhuan Wang.
\newblock {Neural-MCRL}: Neural multimodal contrastive representation learning
  for {EEG}-based visual decoding.
\newblock In \emph{Proceedings of the IEEE International Conference on
  Multimedia and Expo ({ICME})}, pages 1--6, 2025.

\bibitem[Liang et~al.(2020)Liang, Hu, and Feng]{liang2020shot}
Jian Liang, Dapeng Hu, and Jiashi Feng.
\newblock Do we really need to access the source data? {S}ource hypothesis
  transfer for unsupervised domain adaptation.
\newblock In \emph{Proceedings of the International Conference on Machine
  Learning ({ICML})}, pages 6028--6039, 2020.

\bibitem[Lin et~al.(2026)Lin, Wu, and Jung]{lin2026identitytrap}
Jun-You Lin, Ying~Choon Wu, and Tzyy-Ping Jung.
\newblock The identity trap in {EEG} foundation models: A diagnostic audit.
\newblock \emph{arXiv preprint arXiv:2606.06647}, 2026.

\bibitem[Lipton et~al.(2018)Lipton, Wang, and Smola]{lipton2018bbse}
Zachary~C. Lipton, Yu-Xiang Wang, and Alexander~J. Smola.
\newblock Detecting and correcting for label shift with black box predictors.
\newblock In \emph{Proceedings of the International Conference on Machine
  Learning ({ICML})}, pages 3122--3130, 2018.

\bibitem[Lotte et~al.(2018)Lotte, Bougrain, Cichocki, Clerc, Congedo,
  Rakotomamonjy, and Yger]{lotte2018review}
Fabien Lotte, Laurent Bougrain, Andrzej Cichocki, Maureen Clerc, Marco Congedo,
  Alain Rakotomamonjy, and Florian Yger.
\newblock A review of classification algorithms for {EEG}-based brain--computer
  interfaces: a 10 year update.
\newblock \emph{Journal of Neural Engineering}, 15\penalty0 (3):\penalty0
  031005, 2018.

\bibitem[Martin et~al.(2024)Martin, Huang, Shakeri, Pesquet, and
  Ben~Ayed]{martin2024transductive}
S{\'e}gol{\`e}ne Martin, Yunshi Huang, Fereshteh Shakeri, Jean-Christophe
  Pesquet, and Ismail Ben~Ayed.
\newblock Transductive zero-shot and few-shot {CLIP}.
\newblock In \emph{Proceedings of the IEEE/CVF Conference on Computer Vision
  and Pattern Recognition ({CVPR})}, pages 28816--28826, 2024.

\bibitem[Onton et~al.(2006)Onton, Westerfield, Townsend, and
  Makeig]{onton2006imaging}
Julie Onton, Marissa Westerfield, Jeanne Townsend, and Scott Makeig.
\newblock Imaging human {EEG} dynamics using independent component analysis.
\newblock \emph{Neuroscience \& Biobehavioral Reviews}, 30\penalty0
  (6):\penalty0 808--822, 2006.

\bibitem[Oquab et~al.(2024)Oquab, Darcet, Moutakanni, Vo, Szafraniec, Khalidov,
  Fernandez, Haziza, Massa, El-Nouby, Assran, Ballas, Galuba, Howes, Huang, Li,
  Misra, Rabbat, Sharma, Synnaeve, Xu, J{\'e}gou, Mairal, Labatut, Joulin, and
  Bojanowski]{oquab2024dinov2}
Maxime Oquab, Timoth{\'e}e Darcet, Th{\'e}o Moutakanni, Huy~V. Vo, Marc
  Szafraniec, Vasil Khalidov, Pierre Fernandez, Daniel Haziza, Francisco Massa,
  Alaaeldin El-Nouby, Mahmoud Assran, Nicolas Ballas, Wojciech Galuba, Russell
  Howes, Po-Yao Huang, Shang-Wen Li, Ishan Misra, Michael Rabbat, Vasu Sharma,
  Gabriel Synnaeve, Hu Xu, Herv{\'e} J{\'e}gou, Julien Mairal, Patrick Labatut,
  Armand Joulin, and Piotr Bojanowski.
\newblock {DINOv2}: Learning robust visual features without supervision.
\newblock \emph{Transactions on Machine Learning Research}, 2024.

\bibitem[Ozcelik and VanRullen(2023)]{ozcelik2023natural}
Furkan Ozcelik and Rufin VanRullen.
\newblock Natural scene reconstruction from {fMRI} signals using generative
  latent diffusion.
\newblock \emph{Scientific Reports}, 13\penalty0 (1):\penalty0 15666, 2023.

\bibitem[\"{O}zdenizci et~al.(2020)\"{O}zdenizci, Wang, Koike-Akino, and
  Erdo\u{g}mu\c{s}]{ozdenizci2020adversarial}
Ozan \"{O}zdenizci, Ye Wang, Toshiaki Koike-Akino, and Deniz Erdo\u{g}mu\c{s}.
\newblock Learning invariant representations from {EEG} via adversarial
  inference.
\newblock \emph{IEEE Access}, 8:\penalty0 27074--27085, 2020.

\bibitem[Peyr{\'e} and Cuturi(2019)]{peyre2019computational}
Gabriel Peyr{\'e} and Marco Cuturi.
\newblock Computational optimal transport: With applications to data science.
\newblock \emph{Foundations and Trends in Machine Learning}, 11\penalty0
  (5-6):\penalty0 355--607, 2019.

\bibitem[Radford et~al.(2021)Radford, Kim, Hallacy, Ramesh, Goh, Agarwal,
  Sastry, Askell, Mishkin, Clark, Krueger, and Sutskever]{radford2021clip}
Alec Radford, Jong~Wook Kim, Chris Hallacy, Aditya Ramesh, Gabriel Goh,
  Sandhini Agarwal, Girish Sastry, Amanda Askell, Pamela Mishkin, Jack Clark,
  Gretchen Krueger, and Ilya Sutskever.
\newblock Learning transferable visual models from natural language
  supervision.
\newblock In \emph{Proceedings of the International Conference on Machine
  Learning ({ICML})}, pages 8748--8763, 2021.

\bibitem[Radovanovi\'{c} et~al.(2010)Radovanovi\'{c}, Nanopoulos, and
  Ivanovi\'{c}]{radovanovic2010hubs}
Milo\v{s} Radovanovi\'{c}, Alexandros Nanopoulos, and Mirjana Ivanovi\'{c}.
\newblock Hubs in space: Popular nearest neighbors in high-dimensional data.
\newblock \emph{Journal of Machine Learning Research}, 11:\penalty0 2487--2531,
  2010.

\bibitem[Rodrigues et~al.(2019)Rodrigues, Jutten, and
  Congedo]{rodrigues2019rpa}
Pedro Luiz~Coelho Rodrigues, Christian Jutten, and Marco Congedo.
\newblock {R}iemannian {P}rocrustes analysis: Transfer learning for
  brain--computer interfaces.
\newblock \emph{IEEE Transactions on Biomedical Engineering}, 66\penalty0
  (8):\penalty0 2390--2401, 2019.

\bibitem[Samek et~al.(2013)Samek, Meinecke, and
  M\"{u}ller]{samek2013transferring}
Wojciech Samek, Frank~C. Meinecke, and Klaus-Robert M\"{u}ller.
\newblock Transferring subspaces between subjects in brain--computer
  interfacing.
\newblock \emph{IEEE Transactions on Biomedical Engineering}, 60\penalty0
  (8):\penalty0 2289--2298, 2013.

\bibitem[Sch\"{o}nemann(1966)]{schonemann1966procrustes}
Peter~H. Sch\"{o}nemann.
\newblock A generalized solution of the orthogonal {Procrustes} problem.
\newblock \emph{Psychometrika}, 31\penalty0 (1):\penalty0 1--10, 1966.

\bibitem[Scotti et~al.(2023)Scotti, Banerjee, Goode, Shabalin, Nguyen, Cohen,
  Dempster, Verlinde, Yundler, Weisberg, Norman, and
  Abraham]{scotti2023mindeye}
Paul~S. Scotti, Atmadeep Banerjee, Jimmie Goode, Stepan Shabalin, Alex Nguyen,
  Ethan Cohen, Aidan~J. Dempster, Nathalie Verlinde, Elad Yundler, David
  Weisberg, Kenneth~A. Norman, and Tanishq~Mathew Abraham.
\newblock Reconstructing the mind's eye: {fMRI-to-image} with contrastive
  learning and diffusion priors.
\newblock In \emph{Advances in Neural Information Processing Systems
  ({NeurIPS})}, pages 24705--24728, 2023.

\bibitem[Scotti et~al.(2024)Scotti, Tripathy, {Torrico Villanueva}, Kneeland,
  Chen, Narang, Santhirasegaran, Xu, Naselaris, Norman, and
  Abraham]{scotti2024mindeye2}
Paul~S. Scotti, Mihir Tripathy, Cesar~Kadir {Torrico Villanueva}, Reese
  Kneeland, Tong Chen, Ashutosh Narang, Charan Santhirasegaran, Jonathan Xu,
  Thomas Naselaris, Kenneth~A. Norman, and Tanishq~Mathew Abraham.
\newblock {MindEye2}: Shared-subject models enable {fMRI}-to-image with 1 hour
  of data.
\newblock In \emph{Proceedings of the International Conference on Machine
  Learning ({ICML})}, pages 44038--44059, 2024.

\bibitem[She et~al.(2023)She, Zhang, Fang, Ma, and Zhang]{she2023multisource}
Qingshan She, Chenqi Zhang, Feng Fang, Yuliang Ma, and Yingchun Zhang.
\newblock Multisource associate domain adaptation for cross-subject and
  cross-session {EEG} emotion recognition.
\newblock \emph{IEEE Transactions on Instrumentation and Measurement},
  72:\penalty0 1--12, 2023.

\bibitem[Sinkhorn(1964)]{sinkhorn1964}
Richard Sinkhorn.
\newblock A relationship between arbitrary positive matrices and doubly
  stochastic matrices.
\newblock \emph{The Annals of Mathematical Statistics}, 35\penalty0
  (2):\penalty0 876--879, 1964.

\bibitem[Song et~al.(2023)Song, Zheng, Liu, and Gao]{song2023eegconformer}
Yonghao Song, Qingqing Zheng, Bingchuan Liu, and Xiaorong Gao.
\newblock {EEG Conformer}: Convolutional transformer for {EEG} decoding and
  visualization.
\newblock \emph{IEEE Transactions on Neural Systems and Rehabilitation
  Engineering}, 31:\penalty0 710--719, 2023.

\bibitem[Song et~al.(2024)Song, Liu, Li, Shi, Wang, and Gao]{song2024nice}
Yonghao Song, Bingchuan Liu, Xiang Li, Nanlin Shi, Yijun Wang, and Xiaorong
  Gao.
\newblock Decoding natural images from {EEG} for object recognition.
\newblock In \emph{Proceedings of the International Conference on Learning
  Representations ({ICLR})}, 2024.

\bibitem[Spampinato et~al.(2017)Spampinato, Palazzo, Kavasidis, Giordano,
  Souly, and Shah]{spampinato2017deep}
Concetto Spampinato, Simone Palazzo, Isaak Kavasidis, Daniela Giordano, Nasim
  Souly, and Mubarak Shah.
\newblock Deep learning human mind for automated visual classification.
\newblock In \emph{Proceedings of the IEEE Conference on Computer Vision and
  Pattern Recognition ({CVPR})}, pages 6809--6817, 2017.

\bibitem[Takagi and Nishimoto(2023)]{takagi2023highres}
Yu Takagi and Shinji Nishimoto.
\newblock High-resolution image reconstruction with latent diffusion models
  from human brain activity.
\newblock In \emph{Proceedings of the IEEE/CVF Conference on Computer Vision
  and Pattern Recognition ({CVPR})}, pages 14453--14463, 2023.

\bibitem[van~den Oord et~al.(2018)van~den Oord, Li, and Vinyals]{oord2019cpc}
Aaron van~den Oord, Yazhe Li, and Oriol Vinyals.
\newblock Representation learning with contrastive predictive coding.
\newblock \emph{arXiv preprint arXiv:1807.03748}, 2018.

\bibitem[Veilleux et~al.(2021)Veilleux, Boudiaf, Piantanida, and
  Ben~Ayed]{veilleux2021realistic}
Olivier Veilleux, Malik Boudiaf, Pablo Piantanida, and Ismail Ben~Ayed.
\newblock Realistic evaluation of transductive few-shot learning.
\newblock In \emph{Advances in Neural Information Processing Systems
  ({NeurIPS})}, pages 9290--9302, 2021.

\bibitem[Wang et~al.(2021)Wang, Shelhamer, Liu, Olshausen, and
  Darrell]{wang2021tent}
Dequan Wang, Evan Shelhamer, Shaoteng Liu, Bruno Olshausen, and Trevor Darrell.
\newblock Tent: Fully test-time adaptation by entropy minimization.
\newblock In \emph{Proceedings of the International Conference on Learning
  Representations ({ICLR})}, 2021.

\bibitem[Wang et~al.(2022)Wang, Fink, Van~Gool, and Dai]{wang2022cotta}
Qin Wang, Olga Fink, Luc Van~Gool, and Dengxin Dai.
\newblock Continual test-time domain adaptation.
\newblock In \emph{Proceedings of the IEEE/CVF Conference on Computer Vision
  and Pattern Recognition ({CVPR})}, pages 7201--7211, 2022.

\bibitem[Wu et~al.(2025)Wu, Li, Zhang, He, and Ying]{wu2025ubp}
Haitao Wu, Qing Li, Changqing Zhang, Zhen He, and Xiaomin Ying.
\newblock Bridging the vision-brain gap with an uncertainty-aware blur prior.
\newblock In \emph{Proceedings of the IEEE/CVF Conference on Computer Vision
  and Pattern Recognition ({CVPR})}, pages 2246--2257, 2025.

\bibitem[Xu et~al.(2025)Xu, Nunes, Jiang, Ryther, Pringle, Scotti, Delorme, and
  Kneeland]{xu2025alljoined}
Jonathan Xu, Ugo~Bruzadin Nunes, Wangshu Jiang, Samuel Ryther, Jordan Pringle,
  Paul~S. Scotti, Arnaud Delorme, and Reese Kneeland.
\newblock {Alljoined-1.6M}: A million-trial {EEG}-image dataset for evaluating
  affordable brain-computer interfaces.
\newblock \emph{arXiv preprint arXiv:2508.18571}, 2025.

\bibitem[Zanella et~al.(2024)Zanella, G{\'e}rin, and
  Ben~Ayed]{zanella2024transclip}
Maxime Zanella, Beno{\^\i}t G{\'e}rin, and Ismail Ben~Ayed.
\newblock Boosting vision-language models with transduction.
\newblock In \emph{Advances in Neural Information Processing Systems
  ({NeurIPS})}, pages 62223--62256, 2024.

\bibitem[Zanini et~al.(2018)Zanini, Congedo, Jutten, Said, and
  Berthoumieu]{zanini2018transfer}
Paolo Zanini, Marco Congedo, Christian Jutten, Salem Said, and Yannick
  Berthoumieu.
\newblock Transfer learning: A {R}iemannian geometry framework with
  applications to brain--computer interfaces.
\newblock \emph{IEEE Transactions on Biomedical Engineering}, 65\penalty0
  (5):\penalty0 1107--1116, 2018.

\bibitem[Zhang et~al.(2026)Zhang, Wang, Su, Li, Zhang, and
  Zhong]{zhang2025neurobridge}
Wenjiang Zhang, Sifeng Wang, Yuwei Su, Xinyu Li, Chen Zhang, and Suyu Zhong.
\newblock {NeuroBridge}: Bio-inspired self-supervised {EEG}-to-image decoding
  via cognitive priors and bidirectional semantic alignment.
\newblock In \emph{Proceedings of the AAAI Conference on Artificial
  Intelligence}, pages 18028--18036, 2026.

\bibitem[Zhao et~al.(2023)Zhao, Liu, Alahi, and Lin]{zhao2023pitfalls}
Hao Zhao, Yuejiang Liu, Alexandre Alahi, and Tao Lin.
\newblock On pitfalls of test-time adaptation.
\newblock In \emph{Proceedings of the International Conference on Machine
  Learning ({ICML})}, pages 42058--42080, 2023.

\end{thebibliography}
}

\clearpage
\makeatletter
\twocolumn[{%
  \centering
  \includegraphics[width=\textwidth]{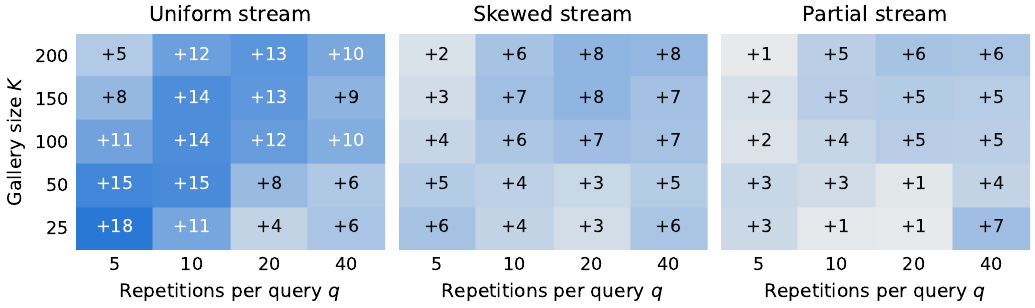}
  \def\@captype{figure}%
  \caption{Margin of GALA over the best other method on the last 30\% of the queries of each stream, in points of top-1 accuracy, for every gallery size $K$ and number of repetitions per query $q$. Blue: GALA is ahead.}
  \label{fig:grid-heatmap}
  \vspace{1.2em}
}]
\makeatother
\appendix
\section{Incremental Results per Gallery Size and Number of Repetitions}
\label{app:stream}

\begin{figure*}[t]
  \centering
  \includegraphics[width=0.72\textwidth]{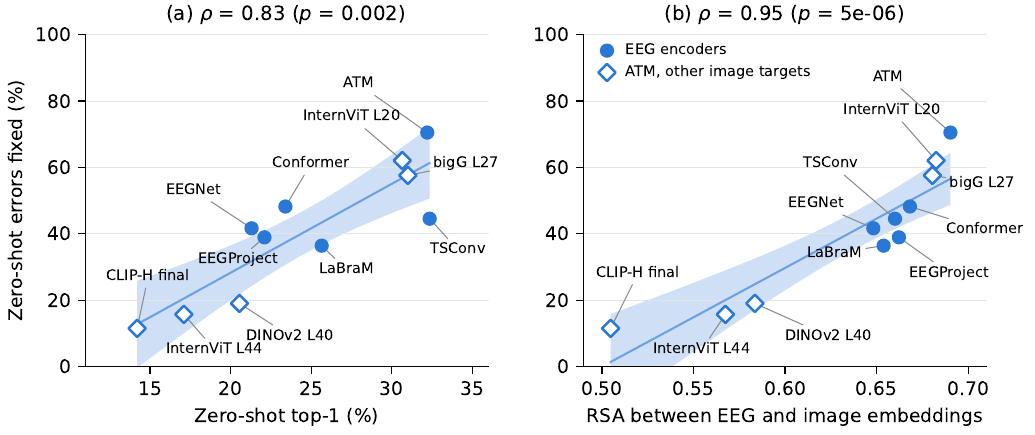}
  \caption{Calibration gain across encoders and image targets in the batch evaluation. Fraction of the zero-shot errors fixed by GALA against (a) the zero-shot top-1 accuracy and (b) the RSA between EEG and image embeddings, for the six encoders of Table~\ref{tab:encoders} (circles) and for ATM trained against five other image targets (diamonds).}
  \label{fig:encoders}
\end{figure*}

\begin{figure*}[p]
  \centering
  \includegraphics[width=\textwidth]{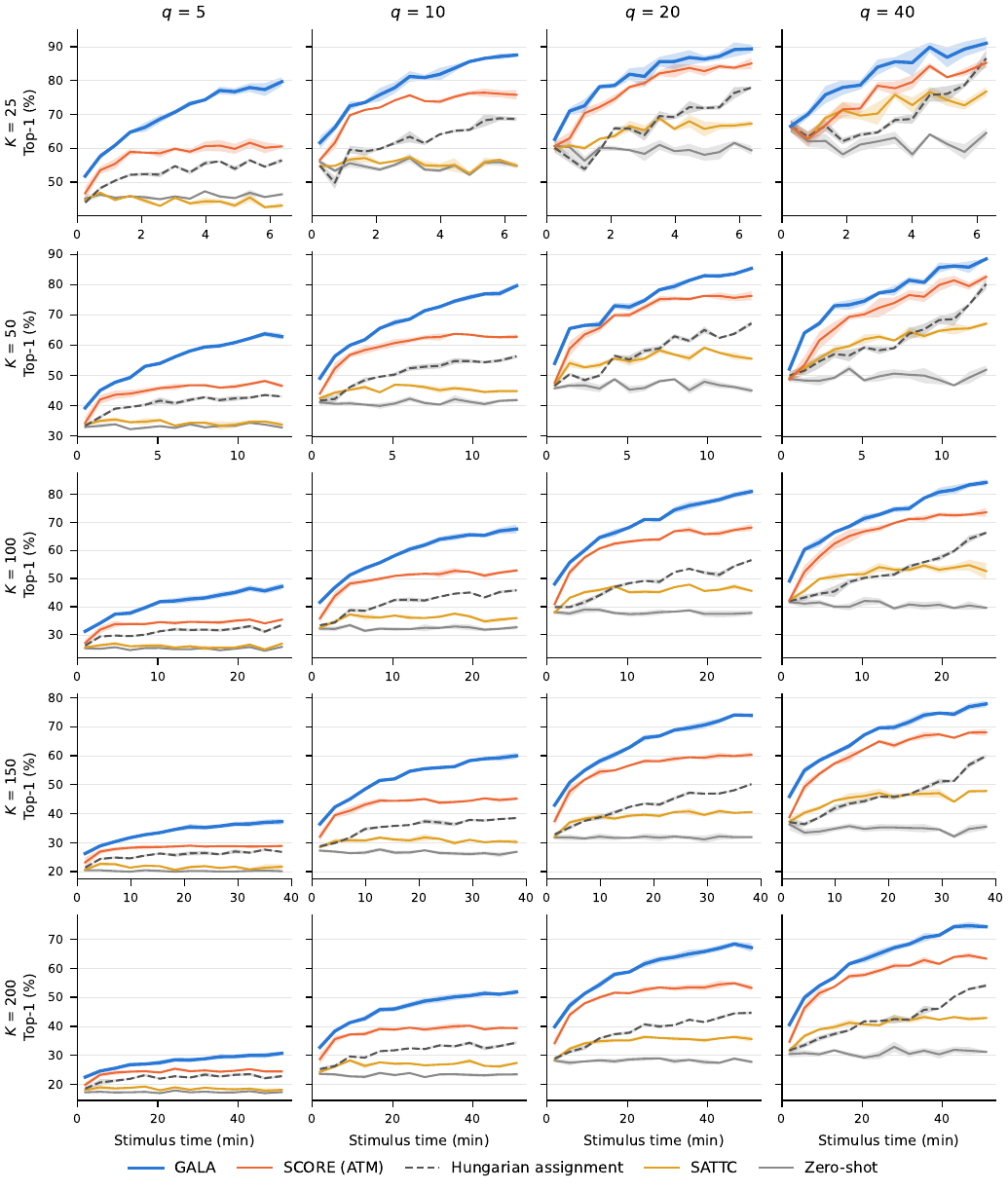}
  \caption{Uniform stream for each gallery size $K$ (rows) and number of averaged repetitions $q$ (columns). Top-1 accuracy of the incoming queries, in windows of one fifteenth of the stream, against stimulus time, at 200\,ms per repetition. }
  \label{fig:full-uniform}
\end{figure*}

\begin{figure*}[p]
  \centering
  \includegraphics[width=\textwidth]{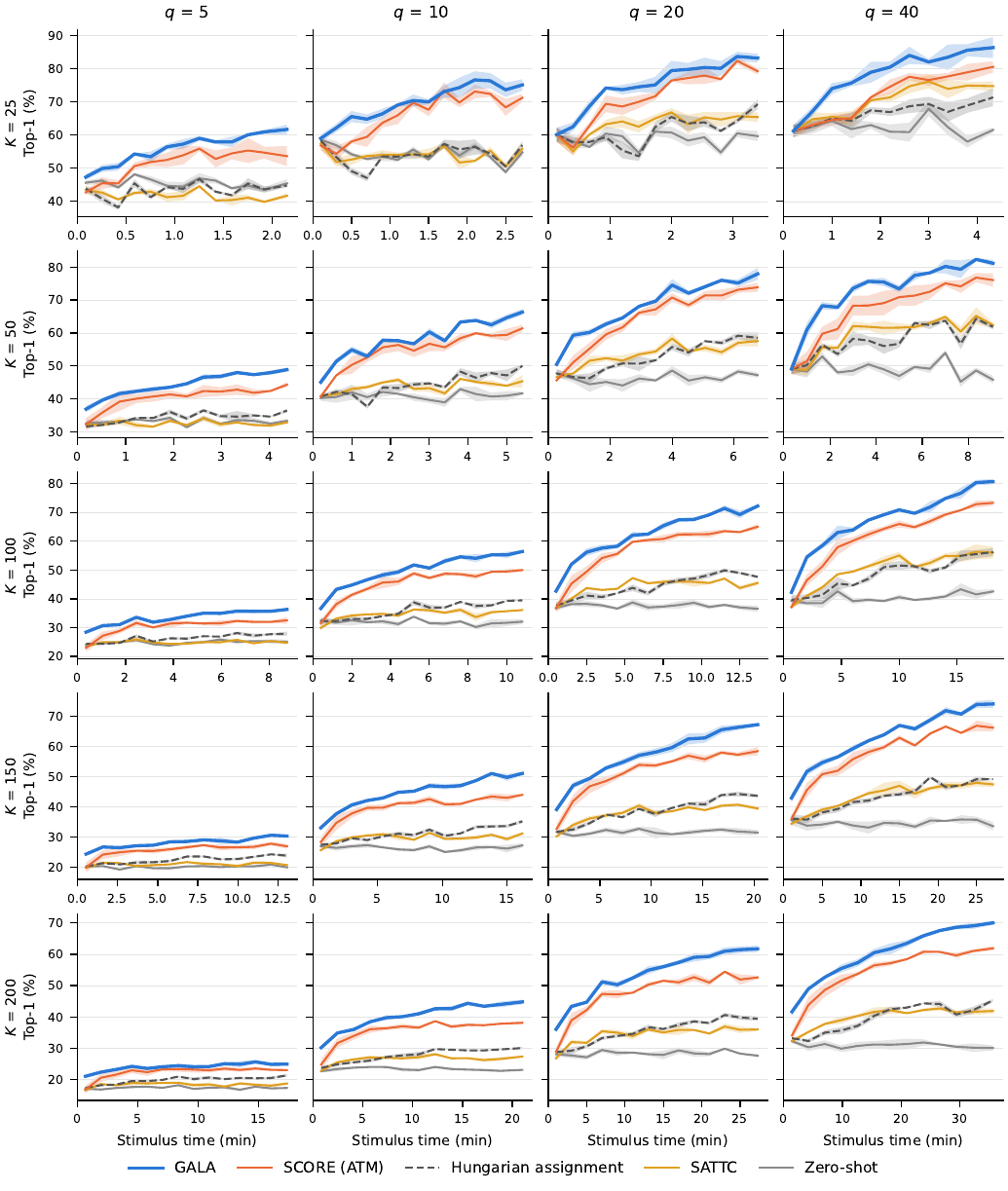}
  \caption{Skewed stream for each gallery size $K$ (rows) and number of averaged repetitions $q$ (columns), as in Fig.~\ref{fig:full-uniform}.}
  \label{fig:full-skew}
\end{figure*}

\begin{figure*}[p]
  \centering
  \includegraphics[width=\textwidth]{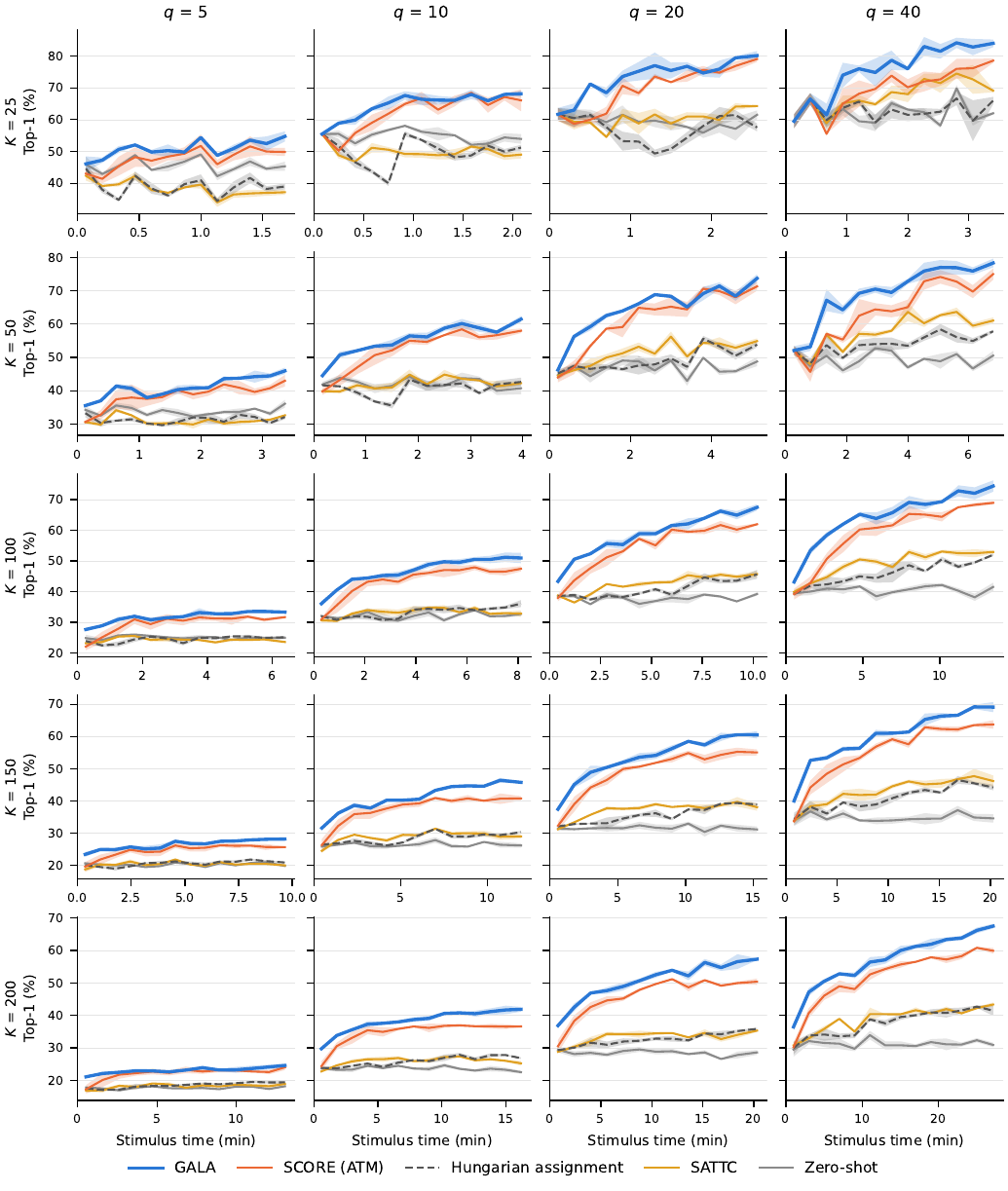}
  \caption{Partial stream for each gallery size $K$ (rows) and number of averaged repetitions $q$ (columns), as in Fig.~\ref{fig:full-uniform}.}
  \label{fig:full-partial}
\end{figure*}

We repeat the incremental protocol of Sec.~\ref{sec:setting} on the uniform, skewed and partial streams, for galleries of $K \in \{25, 50, 100, 150, 200\}$ images drawn at random from the test set and for $q \in \{5, 10, 20, 40\}$ repetitions per query. This extends the results of Sec.~\ref{sec:inc}, which focus on $K=200$ and $q=20$, to smaller galleries and to other numbers of repetitions. We run CSLS, SATTC, the Hungarian assignment, SCORE with ATM and GALA in each configuration, with the hyper-parameters of Sec.~\ref{sec:implementation}, three training seeds and ten stream orders per subject. Figures~\ref{fig:full-uniform}, \ref{fig:full-skew} and~\ref{fig:full-partial} show the accuracy over time on each stream, with a separate y-axis range for each gallery size.

\paragraph{Overview.}
Figure~\ref{fig:grid-heatmap} summarizes the 60 settings, with the accuracy on the last 30\% of the queries of each stream. GALA is the most accurate method in all of them, followed by SCORE with ATM, with an average margin of 6.6 points. The margin is largest on the uniform stream, where every image appears in many queries, and smallest on the partial stream, where some images are never queried and GALA has the least evidence to estimate the query frequencies. GALA is also the most accurate method in all 60 settings for each training seed taken separately. In the main setting, it is above SCORE with ATM for each of the ten subjects, after 20 minutes and over the last 30\% of every stream (two-sided sign test, $p = 0.002$).

\paragraph{Hyper-parameters.}
For each held-out subject, we select $\rho$, $\tau_{\mathrm{sk}}$ and $\gamma$ among 126 combinations, on the streams of the nine other subjects. The 30 folds select neighboring values of the grid. Accuracy also varies little across the grid: on each stream type, the median combination is within about 2 points of the best one over the whole stream.

\section{Base Encoder Training}
\label{app:encoder}

All calibrations act on frozen encoders trained with the same recipe. Previous work~\cite{song2024nice,li2024atm,wu2025ubp,zhang2025neurobridge,du2025shallow} relies on the InfoNCE~\cite{oord2019cpc} loss, in which the EEG embeddings of the same stimulus from different subjects act as false negatives for the image embedding when they fall in the same batch. To prevent this, we use a multi-positive contrastive objective based on stimulus-grouped batches~\cite{khosla2020supcon}: each batch contains $M$ distinct stimuli, each associated with one EEG response from each of the $|\Sc|=9$ source subjects. SubjectMix then replaces each trial by its mixture with the trial of another subject for the same stimulus. For a pair $u=(s,s')$, $s \neq s'$,
\begin{equation}\label{eq:subjmix}
  \xv_{u,i} = \lambda\,\xv_{s,i} + (1 - \lambda)\,\xv_{s',i},
  \quad \lambda \sim \mathrm{Beta}(\alpha_{\mathrm{mix}}, \alpha_{\mathrm{mix}}),
\end{equation}
with $\alpha_{\mathrm{mix}}=0.5$: within a stimulus group, every trial is paired with the trial of a different subject, we draw one $\lambda$ per trial and mix the raw $C\times T$ tensors before the encoder. The image target $\vv_i$ does not change. We index the pairs by $u \in \widetilde{\Sc} = \{(s,s') : s \neq s',\ s,s' \in \Sc\}$ and write $\ev_{u,i} = E_\theta(\xv_{u,i})$. Let $\Uc_i$ be the set of pairs $u$ with a response to stimulus $i$ in the batch, and $\Bc$ the set of all $(u,j)$ in the batch. For a given EEG embedding $\ev_{u,i}$, the image embedding $\vv_i$ is the positive target and the image embeddings $\{\vv_j\}_{j\neq i}$ act as negatives; symmetrically, the image-to-EEG term treats all EEG embeddings of stimulus $i$ as positives:
\begin{align}
\mathcal{L}_{\mathrm{E \to V}} &=
-\frac{1}{M} \sum_{i=1}^{M} \frac{1}{|\Uc_i|}\sum_{u\in\Uc_i}
\log \frac{\exp(\ev_{u,i} \cdot \vv_{i} / \tau)}{\sum_{j=1}^{M} \exp(\ev_{u,i} \cdot \vv_{j} / \tau)}, \\
\mathcal{L}_{\mathrm{V \to E}} &=
-\frac{1}{M} \sum_{i=1}^{M} \frac{1}{|\Uc_i|}\sum_{u\in\Uc_i}
\log \frac{\exp(\vv_{i} \cdot \ev_{u,i} / \tau)}{\sum_{(u',j)\in\Bc} \exp(\vv_{i} \cdot \ev_{u',j} / \tau)},
\end{align}
and the final contrastive loss is $\mathcal{L}_{\mathrm{ctr}} = \frac{1}{2} ( \mathcal{L}_{\mathrm{E \to V}} + \mathcal{L}_{\mathrm{V \to E}} )$. Following recent EEG-to-image alignment practice~\cite{kneeland2025enigma,zhang2025neurobridge}, we $\ell_2$-normalize the image embeddings but keep the EEG embeddings unnormalized during training.

\section{Calibration Gain across Encoders and Image Targets}
\label{app:encoders}

\paragraph{Encoders and image targets.}
In the batch evaluation (Sec.~\ref{sec:conventional}), GALA fixes 36\% to 71\% of the zero-shot errors of the six encoders. The gain of GALA tends to increase with the zero-shot accuracy, with exceptions (Fig.~\ref{fig:encoders}a): ATM and TSConv have roughly the same zero-shot accuracy, but GALA fixes a much larger fraction of the errors of ATM. As an additional diagnostic, we measure the representational similarity (RSA)~\cite{kriegeskorte2008rsa} between the two modalities, as the correlation between the pairwise cosine similarities of the centered test EEG embeddings and those of their images. Two clouds with identical similarity matrices are related by an orthogonal map, so RSA measures how well a rotation, such as the one GALA estimates, can align the EEG cloud with the image cloud. Over the six encoders, ATM has the highest RSA and the largest gain. To widen the range of our experiment, we train ATM against five other image targets: intermediate layers of InternViT-6B (layer 20) and OpenCLIP bigG~\cite{cherti2023openclip} (layer 27), and the final or near-final layers of CLIP ViT-H~\cite{radford2021clip}, InternViT-6B (layer 44) and DINOv2~\cite{oquab2024dinov2} (layer 40). Over the 11 models, RSA predicts the gain better than zero-shot accuracy (Fig.~\ref{fig:encoders}).

\section{Standard Batch Evaluation}
\label{app:batch}

\begin{table*}[t]
\caption{Standard batch evaluation: top-1 accuracy (\%) of 200-way cross-subject retrieval on THINGS-EEG2, per held-out subject, with one 80-repetition query per gallery image.}
\label{tab:main}
\centering
\small
\setlength{\tabcolsep}{4.2pt}
\begin{tabular}{@{}llcccccccccccc@{}}
\toprule
 & Method & S1 & S2 & S3 & S4 & S5 & S6 & S7 & S8 & S9 & S10 & Top-1 & Top-5 \\
\midrule
\multirow{6}{*}{\shortstack[l]{Inductive}}
 & NICE~\cite{song2024nice} & 7.6 & 5.9 & 6.0 & 6.3 & 4.4 & 5.6 & 5.6 & 6.3 & 5.7 & 8.4 & 6.2 & 21.4 \\
 & ATM-S~\cite{li2024atm} & 10.5 & 7.1 & 11.9 & 14.7 & 7.0 & 11.1 & 16.1 & 15.0 & 4.9 & 20.5 & 11.9 & 33.8 \\
 & Neural-MCRL~\cite{li2024mcrl} & 13.0 & 12.0 & 14.5 & 12.5 & 11.5 & 13.5 & 14.0 & 18.5 & 13.5 & 17.0 & 14.0 & 34.3 \\
 & UBP~\cite{wu2025ubp} & 11.5 & 15.5 & 9.8 & 13.0 & 8.8 & 11.7 & 10.2 & 12.2 & 15.5 & 16.0 & 12.4 & 33.4 \\
 & NeuroBridge~\cite{zhang2025neurobridge} & 23.2 & 21.2 & 13.2 & 17.0 & 14.5 & 25.0 & 15.3 & 20.1 & 13.7 & 27.2 & 19.0 & 45.9 \\
 & Shallow Align.~\cite{du2025shallow} & 23.5 & 30.6 & 10.0 & 19.5 & 18.1 & 22.7 & 18.6 & 17.3 & 23.0 & 34.4 & 21.8 & 49.4 \\
\midrule
\multirow{7}{*}{\shortstack[l]{Calibrated}}
 & Zero-shot & 38.5 & 39.7 & 23.7 & 25.5 & 39.8 & 22.7 & 25.5 & 29.2 & 30.8 & 46.7 & 32.2 & 62.5 \\
 & CSLS & 52.0 & 51.8 & 32.2 & 33.7 & 49.0 & 33.3 & 36.8 & 39.7 & 41.8 & 63.2 & 43.4 & 75.5 \\
 & SATTC~\cite{huang2026sattc} & 55.3 & 46.3 & 48.8 & 35.2 & 50.3 & 45.3 & 45.8 & 41.3 & 48.3 & 63.5 & 48.0 & 79.8 \\
 & SCORE~\cite{cui2026score} & 61.5 & 65.0 & 46.0 & 39.0 & 40.0 & 68.5 & 54.0 & 39.0 & 42.5 & 68.0 & 52.4 & 84.0 \\
 & SCORE with ATM & 73.8 & 68.7 & 70.8 & 49.3 & 67.7 & 65.0 & 65.0 & 61.2 & 67.5 & 79.0 & 66.8 & \textbf{92.3} \\
 & Hungarian assignment & 77.0 & 66.8 & 64.0 & 37.8 & 67.5 & 65.0 & 55.0 & 56.0 & 65.2 & 78.8 & 63.3 & 89.5 \\
 \addlinespace[2pt]
 \cdashline{2-14}
 \addlinespace[2pt]
 & GALA & \textbf{88.0} & \textbf{79.0} & \textbf{85.3} & \textbf{56.7} & \textbf{79.3} & \textbf{86.2} & \textbf{77.0} & \textbf{71.0} & \textbf{78.0} & \textbf{90.3} & \textbf{79.1} & 91.9 \\
\bottomrule
\end{tabular}

\vspace{2pt}
\end{table*}

\paragraph{Comparison with published results.}
Published calibration results combine the effect of the calibration with that of the base encoder. SCORE reports 53.2\% with its own encoder, SAMGA~\cite{jiang2026samga}. The code of SCORE is not released, so we train SAMGA with its official code and reimplement the deployment step of SCORE, which reaches 52.4\% (Table~\ref{tab:main}). SATTC reports 14.8\% with its own encoder, whose zero-shot accuracy is 9.2\%, and its official code gives 11.4\% in our runs. We therefore evaluate every calibration method on our main encoder, ATM, and on the five other encoders of Appendix~\ref{app:enc-tables}. In Table~\ref{tab:main}, the inductive rows are published results with their own encoders, and every calibrated row uses ATM, except SCORE, which uses SAMGA. GALA has the best top-1 accuracy for every subject, but SCORE with ATM is 0.4 points above it in top-5 accuracy.

\section{Other Datasets}
\label{app:datasets}

\begin{table*}[t]
\caption{Incremental transductive calibration on Alljoined-1.6M and THINGS-MEG: top-1 accuracy (\%) on the last 30\% of the queries of each stream (ATM, one training seed).}
\label{tab:datasets}
\centering
\small
\setlength{\tabcolsep}{5pt}
\begin{tabular}{@{}lccccccccc@{}}
\toprule
 & \multicolumn{3}{c}{Alljoined-1.6M, $q=20$} & \multicolumn{3}{c}{Alljoined-1.6M, $q=40$} & \multicolumn{3}{c}{THINGS-MEG, $q=6$} \\
\cmidrule(lr){2-4}\cmidrule(lr){5-7}\cmidrule(lr){8-10}
Method & Uniform & Skewed & Partial & Uniform & Skewed & Partial & Uniform & Skewed & Partial \\
\midrule
Zero-shot & 6.9 & 7.1 & 7.1 & 8.7 & 8.7 & 8.4 & 4.6 & 4.7 & 4.6 \\
CSLS & 7.6 & 7.6 & 7.4 & 10.6 & 10.5 & 10.5 & 5.4 & 5.7 & 5.8 \\
SATTC~\cite{huang2026sattc} & 7.3 & 7.5 & 6.7 & 9.8 & 9.5 & 8.8 & 5.2 & 5.1 & 5.7 \\
Hungarian assignment & \textbf{8.6} & 8.6 & 8.1 & 12.2 & 11.7 & 11.0 & \textbf{6.2} & \textbf{7.1} & 6.0 \\
SCORE with ATM & 8.5 & \textbf{9.0} & 8.9 & \textbf{12.5} & 12.2 & 12.3 & 6.0 & 6.2 & 6.7 \\
\midrule
GALA & 8.3 & \textbf{9.0} & \textbf{9.0} & \textbf{12.5} & \textbf{12.4} & \textbf{12.4} & 5.6 & 6.3 & \textbf{7.1} \\
\bottomrule
\end{tabular}
\end{table*}

We run the incremental protocol on two other datasets recorded with the THINGS images. Alljoined-1.6M~\cite{xu2025alljoined} contains EEG recordings of 20 participants with a 32-channel headset, and its test set holds the 200 test images of THINGS-EEG2, with 80 repetitions each. THINGS-MEG~\cite{hebart2023thingsdata} contains MEG recordings of 4 participants with 271 channels, and its test set holds 200 images with 12 repetitions each. On each dataset, we train ATM with the recipe of Sec.~\ref{sec:implementation} under the LOSO protocol, with one seed, at most 50 epochs, early stopping after 5 epochs without improvement of the validation loss, and a batch size of 128 on THINGS-MEG. We select the hyper-parameters of GALA and SCORE on the streams of the other subjects of the same dataset. We use ten stream orders per subject on THINGS-MEG, which has only four subjects, and three on Alljoined-1.6M, which has 20.

Table~\ref{tab:datasets} reports the results. On both datasets, the zero-shot accuracy of ATM is below 10\%, and every label-free method stays within a few points of the zero-shot decoder. GALA is the most accurate method, or close to it, on every stream of Alljoined-1.6M, and on par with the Hungarian assignment and SCORE with ATM on THINGS-MEG. As in Appendix~\ref{app:encoders}, the gain of GALA is small when the zero-shot decoder is weak.

\section{Results across Encoders}
\label{app:enc-tables}

Table~\ref{tab:encoders} reports the incremental results of the six encoders of Sec.~\ref{sec:encoders} on the uniform, skewed and partial streams. After 1 minute, each stream contributes only nine queries, and accuracy varies by a few points between streams. From 5 minutes on, GALA is the most accurate method for every encoder except LaBraM, for which SCORE is ahead after 10 and 20 minutes on every stream, and after 5 minutes on the partial stream.

\section{Ablation of GALA Components}
\label{app:ablation}

Each row of Table~\ref{tab:ablation} changes one component of GALA in the main setting. The uniform stream queries every image equally often, so uniform column masses are the true query frequencies of this stream. This explains why the variant with uniform masses is above GALA at the end of the uniform stream, and below it in all other columns. Without the prior, GALA is below the zero-shot decoder at every reported time, which shows the essential role of the prior in the incremental setting (Sec.~\ref{sec:prior}).

\begin{table*}[p]
\caption{Ablation of GALA components in the incremental setting: top-1 accuracy (\%) after 1 and 20 minutes of each stream, and at the end of the uniform stream.}
\label{tab:ablation}
\centering
\small
\setlength{\tabcolsep}{2.2pt}
\begin{tabular}{@{}lccccccc@{}}
\toprule
 & \multicolumn{3}{c}{Uniform} & \multicolumn{2}{c}{Skewed} & \multicolumn{2}{c}{Partial} \\
\cmidrule(lr){2-4}\cmidrule(lr){5-6}\cmidrule(lr){7-8}
Variant & 1 & 20 & 53 & 1 & 20 & 1 & 20 \\
\midrule
GALA & 37.2 & 59.1 & 67.3 & 37.4 & 59.0 & 37.4 & 56.1 \\
w/o whitening & 37.4 & 55.7 & 60.8 & 37.7 & 55.8 & 37.5 & 53.0 \\
w/o CSLS & 36.2 & 55.8 & 58.6 & 35.4 & 55.3 & 36.4 & 54.5 \\
uniform masses & 37.1 & 58.4 & 69.4 & 37.3 & 58.5 & 37.1 & 55.0 \\
w/o column masses & 37.9 & 58.5 & 65.8 & 37.9 & 58.3 & 37.1 & 55.6 \\
w/o rotation & 36.2 & 39.1 & 39.0 & 34.5 & 39.4 & 34.8 & 37.6 \\
w/o prior & 10.7 & 4.0 & 5.5 & 10.4 & 3.9 & 10.0 & 3.5 \\
16 iterations per update & 37.3 & 59.2 & 67.6 & 37.4 & 59.2 & 37.5 & 56.4 \\
\midrule
Zero-shot & 30.1 & 28.4 & 27.9 & 29.5 & 28.7 & 28.8 & 28.0 \\
\bottomrule
\end{tabular}
\end{table*}

\begin{table*}[p]
\caption{Top-1 accuracy (\%) across encoders after 1, 5, 10 and 20 minutes of the uniform, skewed and partial streams, and at the end of the uniform stream (53 minutes).}
\label{tab:encoders}
\centering
\small
\setlength{\tabcolsep}{4.5pt}
\begin{tabular}{@{}llccccccccccccc@{}}
\toprule
 & & \multicolumn{5}{c}{Uniform} & \multicolumn{4}{c}{Skewed} & \multicolumn{4}{c}{Partial} \\
\cmidrule(lr){3-7}\cmidrule(lr){8-11}\cmidrule(lr){12-15}
Encoder & Method & 1 & 5 & 10 & 20 & 53 & 1 & 5 & 10 & 20 & 1 & 5 & 10 & 20 \\
\midrule
\multirow{4}{*}{ATM} & Zero-shot & 30.1 & 27.8 & 28.0 & 28.4 & 27.9 & 29.5 & 27.4 & 28.8 & 28.7 & 28.8 & 28.2 & 29.1 & 28.0 \\
 & SATTC & 28.2 & 32.0 & 34.2 & 35.5 & 35.9 & 27.5 & 32.2 & 35.1 & 35.7 & 28.3 & 33.2 & 34.3 & 34.3 \\
 & SCORE & 29.3 & 42.3 & 48.2 & 52.0 & 53.4 & 29.3 & 42.2 & 48.1 & 52.1 & 29.9 & 43.9 & 48.8 & 50.1 \\
 & GALA & \textbf{37.2} & \textbf{46.2} & \textbf{51.6} & \textbf{59.1} & \textbf{67.3} & \textbf{37.4} & \textbf{45.3} & \textbf{52.0} & \textbf{59.0} & \textbf{37.4} & \textbf{47.4} & \textbf{51.7} & \textbf{56.1} \\
\midrule
\multirow{4}{*}{TSConv} & Zero-shot & 21.7 & 24.6 & 23.5 & 25.1 & 22.1 & \textbf{30.0} & 22.8 & 25.2 & 24.7 & \textbf{23.3} & 25.6 & 24.8 & 23.4 \\
 & SATTC & 17.0 & 18.4 & 19.5 & 20.7 & 19.2 & 18.0 & 19.0 & 21.2 & 19.9 & 16.2 & 20.5 & 20.6 & 19.9 \\
 & SCORE & 20.0 & 26.9 & 30.2 & 34.6 & 34.2 & 17.3 & 26.6 & 30.7 & 35.0 & 18.8 & 28.1 & 31.3 & 32.4 \\
 & GALA & \textbf{26.2} & \textbf{30.8} & \textbf{34.1} & \textbf{40.9} & \textbf{51.1} & 29.3 & \textbf{30.5} & \textbf{36.9} & \textbf{43.0} & 22.5 & \textbf{33.4} & \textbf{35.8} & \textbf{38.8} \\
\midrule
\multirow{4}{*}{EEGNet} & Zero-shot & 17.9 & 18.3 & 17.2 & 18.4 & 18.5 & 28.0 & 17.2 & 19.2 & 18.7 & 15.8 & 18.4 & 17.9 & 18.5 \\
 & SATTC & 15.4 & 18.1 & 19.2 & 20.8 & 20.9 & 19.3 & 17.9 & 20.9 & 21.7 & 20.4 & 21.5 & 20.6 & 20.2 \\
 & SCORE & 15.0 & 26.9 & 29.4 & 32.6 & 33.2 & 22.0 & 26.2 & 30.8 & 33.5 & 20.8 & 27.3 & 30.3 & 32.0 \\
 & GALA & \textbf{20.8} & \textbf{29.0} & \textbf{31.6} & \textbf{36.4} & \textbf{43.0} & \textbf{32.0} & \textbf{27.0} & \textbf{34.2} & \textbf{37.8} & \textbf{24.6} & \textbf{31.4} & \textbf{33.4} & \textbf{35.9} \\
\midrule
\multirow{4}{*}{EEGProject} & Zero-shot & 18.3 & 20.2 & 17.7 & 19.2 & 18.1 & 22.0 & 19.5 & 20.2 & 18.6 & 17.1 & 18.8 & 19.6 & 18.9 \\
 & SATTC & 20.2 & 18.5 & 18.3 & 20.5 & 18.8 & 18.7 & 19.5 & 20.6 & 19.1 & 18.3 & 19.6 & 19.6 & 19.5 \\
 & SCORE & 16.7 & 27.2 & 27.9 & 30.0 & 28.6 & 16.0 & 27.2 & 30.6 & 30.9 & 19.2 & 28.0 & 29.7 & 29.7 \\
 & GALA & \textbf{24.2} & \textbf{29.8} & \textbf{30.8} & \textbf{33.9} & \textbf{35.7} & \textbf{26.7} & \textbf{29.6} & \textbf{33.7} & \textbf{34.5} & \textbf{20.8} & \textbf{29.4} & \textbf{33.1} & \textbf{33.2} \\
\midrule
\multirow{4}{*}{Conformer} & Zero-shot & 19.2 & 20.2 & 19.5 & 19.7 & 18.8 & 20.0 & 18.9 & 20.5 & 20.6 & 20.8 & 20.2 & 19.3 & 19.8 \\
 & SATTC & 21.8 & 18.0 & 18.5 & 20.6 & 20.2 & 20.0 & 18.9 & 21.3 & 21.2 & 24.6 & 20.5 & 20.0 & 19.9 \\
 & SCORE & 21.2 & 25.3 & 29.2 & 31.8 & 33.3 & 20.7 & 25.2 & 31.1 & 33.3 & 15.4 & 27.7 & 28.8 & 31.4 \\
 & GALA & \textbf{28.3} & \textbf{30.3} & \textbf{33.6} & \textbf{39.7} & \textbf{48.7} & \textbf{34.0} & \textbf{29.5} & \textbf{35.6} & \textbf{39.8} & \textbf{29.6} & \textbf{31.5} & \textbf{32.3} & \textbf{37.1} \\
\midrule
\multirow{4}{*}{LaBraM} & Zero-shot & 25.8 & 21.8 & 20.8 & 22.7 & 21.7 & 27.3 & 22.5 & 23.6 & 21.9 & 20.8 & 22.2 & 23.3 & 23.1 \\
 & SATTC & 21.3 & 25.6 & 27.0 & 29.9 & 29.2 & 22.7 & 25.9 & 29.8 & 29.8 & 21.7 & 28.0 & 29.8 & 29.3 \\
 & SCORE & 20.8 & 30.8 & \textbf{35.0} & \textbf{39.2} & 39.0 & 20.7 & 33.3 & \textbf{38.8} & \textbf{38.8} & 24.2 & \textbf{35.8} & \textbf{35.7} & \textbf{37.7} \\
 & GALA & \textbf{29.2} & \textbf{32.2} & 34.3 & 37.5 & \textbf{40.4} & \textbf{31.3} & \textbf{34.4} & 37.3 & 38.2 & \textbf{28.3} & 34.7 & 34.3 & 36.9 \\
\bottomrule
\end{tabular}
\end{table*}

\end{document}